\documentclass{article}

\PassOptionsToPackage{numbers, compress}{natbib}
\usepackage[final]{nips_2016}

\usepackage[utf8]{inputenc} % allow utf-8 input
\usepackage[T1]{fontenc}    % use 8-bit T1 fonts
\usepackage{hyperref}       % hyperlinks
\usepackage{url}            % simple URL typesetting
\usepackage{booktabs}       % professional-quality tables
\usepackage{amsfonts}       % blackboard math symbols
\usepackage{nicefrac}       % compact symbols for 1/2, etc.
\usepackage{microtype}      % microtypography
\usepackage{mathtools}

\newtheorem{lemma}{Lemma}
\newtheorem{theorem}{Theorem}
\newtheorem{definition}{Definition}
\newtheorem{assumption}{Assumption}

\DeclareMathOperator*{\argmin}{\arg\!\min}

\DeclareMathOperator\conv{conv}

\DeclareMathOperator\interior{int}

\newcommand{\omegasqrt}{\Omega( \sqrt{T} )}
\newcommand{\argminf}{\argmin_{f \in \mathcal{F}}}
\newcommand{\stdgame}{(\ell, \mathcal{Z}, \mathcal{F})}
\newcommand{\tr}{\ensuremath{{\scriptscriptstyle\mathsf{T}}}}
\newcommand{\norm}[1]{\left\|#1\right\|}

\title{No-Regret Learnability for Piecewise Linear Losses}

\author{
  Arthur Flajolet \\
  Operations Research Center\\
  Massachusetts Institute of Technology\\
  Cambridge, MA 02139 \\
  \texttt{flajolet@mit.edu} \\
  \And
  Patrick Jaillet\\
  Dept. of Electrical Engineering and Computer Science \\
  Operations Research Center\\	
  Massachusetts Institute of Technology\\
  Cambridge, MA 02139\\
  \texttt{jaillet@mit.edu} \\
}

\begin{document}

\maketitle

\begin{abstract}
In the convex optimization approach to online regret minimization, many methods have been developed to guarantee a $O(\sqrt{T})$ bound on regret for subdifferentiable convex loss functions with bounded subgradients, by using a reduction to linear loss functions. This suggests that linear loss functions tend to be the hardest ones to learn against, regardless of the underlying decision spaces. We investigate this question in a systematic fashion looking at the interplay between the set of possible moves for both the decision maker and the adversarial environment. This allows us to highlight sharp distinctive behaviors about the learnability of piecewise linear loss functions. On the one hand, when the decision set of the decision maker is a polyhedron, we establish $\Omega(\sqrt{T})$ lower bounds on regret for a large class of piecewise linear loss functions with important applications in online linear optimization, repeated zero-sum Stackelberg games, online prediction with side information, and online two-stage optimization. On the other hand, we exhibit $o(\sqrt{T})$ learning rates, achieved by the Follow-The-Leader algorithm, in online linear optimization when the boundary of the decision maker's decision set is curved and when $0$ does not lie in the convex hull of the environment's decision set. Hence, the curvature of the decision maker’s decision set is a determining factor for the optimal learning rate. These results hold in a completely adversarial setting.
\end{abstract}

\section{Introduction}

Online convex optimization has emerged as a popular approach to online learning, bringing together convex optimization methods to tackle problems where repeated decisions need to be made in an unknown, possibly adversarial, environment. A full-information online convex optimization problem is a repeated zero-sum game between a learner (the player) and the environment (the opponent). There are $T$ time periods. At each round $t$, the player has to choose $f_t$ in a convex set $\mathcal{F}$. Subsequent to the choice of $f_t$, the environment reveals $z_t \in \mathcal{Z}$ and the loss incurred to the player is $\ell(z_t, f_t)$, for a loss function $\ell$ that is convex in its second argument. Both players are aware of all the parameters of the game, namely $\ell$, $\mathcal{Z}$, and $\mathcal{F}$, prior to starting the game. Additionally, at the end of each period, the opponent's move is revealed to the player. The performance of the player is measured in terms of a quantity coined regret, defined as the gap between the accumulated losses incurred by the player and the best performance he could have achieved in hindsight with a non-adaptive strategy:
\begin{equation*}
	r_T((z_t)_{t=1, \cdots, T}, (f_t)_{t=1, \cdots, T}) = \sum_{t=1}^T \ell(z_t, f_t) - \inf_{f \in \mathcal{F}} \sum_{t=1}^T \ell(z_t, f).
\end{equation*}
In this field, one of the primary focus is to design algorithms, i.e., strategies to select $(f_t)_{t=1, \cdots, T}$ so as to keep the regret as small as possible even when facing an adversarial opponent. Particular emphasis is placed on how the regret scales with $T$ because this dependence relates to a notion of learning rate. If $r_T = o(T)$, the player is, in some sense, learning the game in the long-run since the gap between experienced and best achievable average cumulative payoffs vanishes as $T \rightarrow \infty$. Furthermore, the smaller the growth rate of $r_T$, the faster the learning. A natural question to ask is what is the best learning rate that can be achieved for a given game $\stdgame$. Mathematically, this is equivalent to characterizing the growth rate of the smallest regret that can be achieved by a player against a completely adversarial opponent, expressed as:
\begin{equation}
	\label{eq-RegretExpression}
	R_T(\ell, \mathcal{Z}, \mathcal{F}) = \inf\limits_{f_1 \in \mathcal{F}} \sup\limits_{z_1 \in \mathcal{Z}} \cdots \inf\limits_{f_T \in \mathcal{F}} \sup\limits_{z_T \in \mathcal{Z}} [ \; \sum_{t=1}^T \ell(z_t, f_t) - \inf\limits_{f \in \mathcal{F}} \sum_{t=1}^T \ell(z_t, f) \; ],
\end{equation}
which we refer to as the value of the game $\stdgame$. Aside from pure learning considerations, the growth rate of $R_T\stdgame$ has important consequences in a variety of fields where no-regret algorithms are used to compute complex quantities, e.g. Nash equilibria in Game Theory \cite{nisan2007algorithmic} or solutions to optimization problems in convex optimization \cite{freund1999adaptive}, in which case this growth rate translates into the number of iterations required to compute the quantity with a given precision. We investigate this question in a systematic fashion by looking at the interplay between $\mathcal{F}$ and $\mathcal{Z}$ for the following class of piecewise linear loss functions:
\begin{equation}
	\label{eq-def-general-loss}
	\ell(z, f) = \max_{ x \in \mathcal{X}(z) } (C(z)f + c(z))^\tr x,
\end{equation}	
where, for any $z \in \mathcal{Z}$, $C(z)$ is a matrix, $c(z)$ is a vector, and $\mathcal{X}(z) \subset \mathbb{R}^d$ is either a finite set or a polyhedron $\{x \in \mathbb{R}^d \; | \; A(z) x \leq b(z)\}$ with $A(z)$ a matrix and $b(z)$ a vector. This type of loss functions arises in a number of important contexts such as online linear optimization, repeated zero-sum Stackelberg games, online prediction with side information, and online two-stage optimization, as illustrated in Section \ref{sec-applications}. Throughout the paper, we make the following standard assumption so as to have the game well-defined.
\begin{assumption}
\label{assumption-sets-all}
	$\mathcal{Z}$ is a non-empty compact subset of $\mathbb{R}^{d_z}$ and $\mathcal{F}$ is a non-empty, convex, and compact subset of $\mathbb{R}^{d_f}$. For any choice of $z \in \mathcal{Z}$, the set $\mathcal{X}(z)$ is not empty. The loss function $\ell$ is bounded on $\mathcal{Z} \times \mathcal{F}$. Moreover, either $\mathcal{Z}$ has finite cardinality or $\ell(\cdot, f)$ is continuous for any $f \in \mathcal{F}$.
\end{assumption}

\paragraph{Contributions.}
A number of no-regret algorithms developed in the literature can be used as a black box for the settings considered in this paper in order to get $O(\sqrt{T})$ bounds on regret, e.g. Exponential Weights \cite{vovk1990aggregating}, Online Gradient Descent \cite{zinkevich2003online}, and more generally Online Mirror Descent \cite{herbster1998tracking}, to cite a few. To get better learning rates, other approaches have been proposed but they usually rely on either the curvature of $\ell$, for instance if $\ell$ is strongly convex in its second argument \cite{hazan2007logarithmic}, which is not the case here, or more information about the sequence $(z_t)_{t=1, \cdots, T}$, see for example \cite{rakhlin2013optimization}, which is not available in the fully adversarial setting. Aside from particular instances, e.g. \cite{cesa2006prediction} and \cite{abernethy2008optimal}, it is in general unknown how the interplay between $\ell$, $\mathcal{Z}$, and $\mathcal{F}$ determines the growth rate of $R_T(\ell, \mathcal{Z}, \mathcal{F})$. The main insight from this paper is that the curvature of the decision maker’s set of moves is a determining factor for the growth rate of $R_T(\ell, \mathcal{Z}, \mathcal{F})$: if $\mathcal{F}$ has rough edges then we are doomed to a rate of $\Theta(\sqrt{T})$, otherwise, if it is curved, the rate can be exponentially smaller. Specifically, we show that:
\begin{enumerate}
	\item When $\mathcal{F}$ is a polyhedron, either $R_T\stdgame = 0$ or $R_T\stdgame = \omegasqrt$. This lower bound applies to online combinatorial optimization where $\mathcal{F}$ is a combinatorial set, to many experts settings and repeated zero-sum Stackelberg games where the player resorts to a randomized strategy, as well as to many online prediction problems with side information and online two-stage optimization problems. 
	\item When (i) $\ell$ is linear, (ii) $\mathcal{F} = \{f \in \mathbb{R}^{d_f} \; | \; F(f) \leq 0 \}$, for $F$ either a strongly convex function or $F(f) = \norm{f}_\mathcal{F} - C$ where $C \geq 0$ and $\norm{ \; }_\mathcal{F}$ is a $q-$uniformly convex norm with $q \in [2, 3]$, and (iii) $0$ does not lie in the convex hull of $\mathcal{Z}$, we have $R_T\stdgame = o(\sqrt{T})$, achieved by the Follow-The-Leader algorithm \cite{kalai2005efficient}. This result applies to repeated zero-sum games where the player picks a cost vector (e.g. arc costs) of bounded euclidean norm and the opponent chooses an element in a combinatorial set (e.g. a path). This also applies to non-linear loss functions when $0$ does not lie in the convex hull of the set of subgradients of $\ell$ with respect to the second-coordinate by a standard reduction to linear loss functions, see \cite{zinkevich2003online}. Note that assumption (iii) is required to get $o(\sqrt{T})$ rates as it is well known that $R_T\stdgame = \omegasqrt$ for linear losses when $0$ lies in the interior of the convex hull of $\mathcal{Z}$, see Section \ref{sec-lower-bound}.
\end{enumerate}

\subsection{Applications}
\label{sec-applications}
We list examples of situations where losses of the type \eqref{eq-def-general-loss} arise.

\paragraph{Online linear optimization}
In this setting, the loss function is given by $\ell(z, f) = z^\tr f$. This includes, in particular:
\begin{itemize}
	\item online combinatorial optimization where the opponent picks a cost in $[0, 1]^{d_z}$ and $\mathcal{F}$ is defined as the convex hull of a finite set of elements (e.g. paths, spanning trees, and matchings),	
	\item experts settings where the player picks a distribution over the experts' advice (in which case $\mathcal{F}$ is also a polyhedron) and the opponent reveals a cost for each of the experts.
\end{itemize}
In online linear optimization, regret lower bounds are often derived by introducing a randomized zero-mean i.i.d. opponent, see \cite{abernethy2008optimal}. However, this is possible only if $0$ is in interior of the convex hull of $\mathcal{Z}$, which is typically not the case in online combinatorial optimization. A general feature of online linear optimization that will turn out to be important in the analysis is that there is no loss of generality in assuming that $\mathcal{Z}$ is a convex set in the following sense.
\begin{lemma}
\label{lemma-LinearConvexGameIsEquivalent}
When $\ell(z, f) = z^\tr f$, the games $(\ell, \mathcal{Z}, \mathcal{F})$ and $(\ell, \conv( \mathcal{Z} ), \mathcal{F})$ are equivalent, i.e.:
\begin{equation*}
	R_T(\ell, \mathcal{Z}, \mathcal{F}) = R_T(\ell, \conv( \mathcal{Z} ), \mathcal{F}).
\end{equation*}
\end{lemma}

\paragraph{Repeated zero-sum Stackelberg games} A repeated zero-sum Stackelberg game is a repeated zero-sum game with the particularity that one of the players, referred to as the leader, has to commit first to a randomized strategy $f$ without even knowing which of the $N$ other players, indexed by $z$, he is going to face at the next round. The interaction between the leader and player $z \in \{1, \cdots, N\}$ is captured by a payoff matrix $M(z)$. Once the leader is set on a strategy, the identity of the other player is revealed and the latter best-responds to the leader's strategy, leading to the following expression for the loss function: 
\begin{equation*}
	\ell(z, f) = \max_{i = 1, \cdots, I_z} e_i^\tr M(z) f,
\end{equation*}
where $I_z$ is the number of possible moves for player $z$. We illustrate with a network security problem that has applications in  urban network security \cite{jain2013security} and fare evasion prevention in transit networks \cite{jiang2012towards}. Consider a directed graph $G = (V, E)$. The leader has a limited number of patrols that can be assigned to arcs in order to intercept the attackers. A configuration $\gamma \in \Gamma$ corresponds to a valid assignment of patrols to arcs and is represented by a vector $(Y^\gamma_{ij})_{(i, j) \in E}$ with $Y^\gamma_{ij} = 1$ if a patrol is assigned to arc $(i, j)$ and $Y^\gamma_{ij} = 0$ otherwise. The leader chooses a mixed strategy $f$ over the set of feasible allocations. Attacker $z \in \{(i_1, j_1), \cdots, (i_N, j_N)\}$ wants to go from $z_1$ to $z_2$ while minimizing the probability of being intercepted. This interaction is captured by the loss function:
\begin{equation*}
		\ell(z, f) = \max_{x \in \mathcal{X}(z)} \sum_{\gamma \in \Gamma} - f_\gamma x_\gamma,
\end{equation*}
with:
\begin{equation*}
	\mathcal{X}(z) = \{ (\max_{(i, j) \in E} X^\pi_{ij} Y^\gamma_{ij} )_{\gamma \in \Gamma} \; | \; \pi \in \Pi(z) \},
\end{equation*}
where $\Pi(z)$ is the set of directed paths joining $z_1$ to $z_2$ in $G$ and $X^\pi_{ij} = 1$ if $(i, j) \in \pi$ and $X^\pi_{ij} = 0$ otherwise. The presentation of repeated Stackelberg games given here follows the model introduced by \citet{balcan2015commitment} for general, i.e. not necessarily zero-sum, Stackelberg security games. In this more general setting, the loss function may not be convex and a possible approach, see \cite{balcan2015commitment}, is to add another layer of randomization which casts the problem back into the realm of online linear optimization.

\paragraph{Online prediction with side-information} This setting has a slightly different flavor as the opponent provides some side information $x$ before the player gets to pick $f \in \mathcal{F}$, subsequent to what the opponent reveals the correct prediction $y$. Nonetheless, the lower bounds established in this paper also apply to this setting through a reduction to the setting without side-information, as detailed at the end of Section \ref{sec-lower-bound}. In the standard linear binary prediction problem, where $\mathcal{F}$ is a $L^2$ ball, $y \in \{-1, 1\}$, and $x$ lies in a $L^2$ ball, loss functions of the form \eqref{eq-def-general-loss} are commonly used, e.g. the absolute loss $\ell((x, y), f) = | y - x^\tr f |$ and the hinge loss $\ell((x, y), f) = \max(0, 1 - y x^\tr f )$. This is also true for linear multiclass prediction problems with the multiclass hinge loss:
\begin{equation*}
	\ell(f, (x, y)) = \max_{j=1, \cdots, N} ( 1\{ j \neq y \} + f_j^\tr x - f_y^\tr x),
\end{equation*}
where $N$ denotes the number of classes, $y \in \{1, \cdots, N\}$, and $f$ is a vector obtained by concatenation of the vectors $f_1, \cdots, f_N$. In the online approach to collaborative filtering, a typical loss function is $\ell(M, (i, j, y)) = |M(i, j) - y|$ where $M$ is a (user, item) matrix with bounded trace norm, $(i, j)$ is a (user, item) pair, and $y$ is the rating of item $j$ by user $i$.

\paragraph{Online two-stage optimization} This setting captures situations where the decision making process can be broken down into two consecutive stages. In the first stage, the player makes a decision represented by $f \in \mathcal{F}$. Subsequently, the opponent discloses some information $z \in \mathcal{Z}$, e.g. a demand vector, and then the player chooses another decision vector $x$ in the second stage, taking into account this newly available information to optimize his objective function. The loss function takes on the following form:
\begin{equation*}
	\ell(z, f) = c_1^\tr f +  \min_{ \substack{ x \in \mathbb{R}^d \\ A f + B x \leq z } } c_2^\tr x,
\end{equation*}
where $c_1$ and $c_2$ are cost vectors and $A$ and $B$ are matrices. Using strong duality, this loss function can be expressed in the canonical form \eqref{eq-def-general-loss}. This framework finds applications in the operation of power grids, where $z$ represents the demand in electricity or the availability of various energy sources. Since $z$ is unknown when it is time to set up conventional generators, the decision maker has to adjust the production or buy additional capacity from a spot market to meed the demand, see for example \cite{kim2014online}.

\paragraph{Congestion control}
We consider a variant of the congestion network game described in \cite{bonifaci2010stackelberg}. A decision maker has to decide how to ship a given set commodities through a network $G = (V, E)$. His decision can be equivalently represented by a flow vector $f$. Because the amount of commodities is assumed to be substantial, implementing $f$ will cause congestion which will impact the other users of the network, represented by a flow vector $z$. The problem faced by the decision maker is to cause as little delay as possible to the other users with the additional difficulty that the traffic pattern $z$ is not known ahead of time. Each arc $e \in E$ has an associated latency function that is convex in the flow on this arc:
\begin{equation*}
	c_e(f + z) = \max_{k=1, \cdots, K} ( c^k_e \cdot (f_e + z_e) + s^k_e ),
\end{equation*}
As a result, the total delay incurred to the other users can be expressed as:
\begin{equation*}
	\ell(z, f) = \sum_{e \in E} z_{e} \max_{k=1, \cdots, K} ( c^k_e \cdot (f_e + z_e) + s^k_e ).
\end{equation*}

\subsection{Related work}
Asymptotically matching lower and upper bounds on $R_T\stdgame$ can be found in the literature for a variety of loss functions although the discussion tends to be restrictive as far as the decision sets $\mathcal{F}$ and $\mathcal{Z}$ are concerned. The value of the game is shown to be $\Theta( \log{T} )$ for three standard examples of curved loss functions. The first example, studied by \citet{abernethy2008optimal}, is the quadratic loss where $\ell(z, f) = z \cdot f + \sigma \| f \|_2^2$ for $\sigma > 0$, with $\mathcal{Z}$ and $\mathcal{F}$ bounded $L^2$ balls. The second, studied by \citet{vovk1998competitive}, is the online linear regression setting where the opponent plays $z = (y, x) \in \mathcal{Z} = [-C_y, C_y] \times B_{\infty}(0, 1)$ for $C_{y} > 0$ ($B_{\infty}(0, 1)$ denotes the unit $L^\infty$ ball), the loss is $\ell((y, x), f) = (y - x \cdot f)^2$, and $\mathcal{F}$ is an $L^2$ ball. The last one, from \citet{ordentlich1998cost}, is the log-loss $\ell(z, f) = - \log{z \cdot f}$ with $\mathcal{Z}$ any compact set in $\mathbb{R}^d$ and $\mathcal{F}$ the simplex of dimension $d$. For non-curved losses, evidence suggests that the value of the game increases exponentially to $\Theta(\sqrt{T})$. Indeed, $\omegasqrt$ lower bounds are proved for several instances involving the absolute loss $\ell(z, f) = | z - f |$ in \cite{cesa1999prediction}, typically with $\mathcal{Z} = \{ 0, 1 \}$ and $\mathcal{F} = [0, 1]$. For purely linear loss functions, \citet{abernethy2008optimal} establish a $\Omega( \sqrt{T} )$ lower bound on $R_T\stdgame$ when $\mathcal{Z}$ is an $L^2$ ball centered at $0$ and $\mathcal{F}$ is either an $L^2$ ball or a bounded rectangle. This result was later generalized in \cite{abernethy2009stochastic} and shown to hold for $\mathcal{F}$ a unit ball in any norm centered at $0$ and $\mathcal{Z}$ its dual ball. \citet{cesa2006prediction} investigate the experts setting, i.e. $\mathcal{Z} = [0,1]^d$ and $\mathcal{F}$ is the simplex in dimension $d$, and proves the same $\omegasqrt$ lower bound (which also holds if $\mathcal{Z}$ is the simplex in dimension $d$, see \cite{abernethy2009stochastic}). \citet{rakhlin2015online} establish $\omegasqrt$ lower bounds on regret when $\ell$ is the absolute loss for a prediction with side-information setting more general than the one considered in this paper where the player picks a function $f(\cdot)$, the opponent picks a pair $(x, y)$, and the loss is $\ell(f(\cdot), (x, y)) = | f(x) - y|$. The list of results listed above is far from being exhaustive but provides a good picture of the current state of the art. For each loss function, the intrinsic limitations of online algorithms are well-understood, usually with the construction of a particular example of $\mathcal{F}$ and $\mathcal{Z}$ for which a lower bound on $R_T\stdgame$ asymptotically matches the best guarantee achieved by one of these algorithms. We aim at studying lower bounds on the value of the game in a more systematic fashion using tools rooted in duality theory and sensitivity analysis. All the proofs are deferred to the Appendix.

\paragraph{Notations}
For a set $S \subset \mathbb{R}^d$, $\conv( S )$ (resp. $\interior( S )$) refers to the convex hull (resp interior) of this set. When $S$ is a compact, we define $\mathcal{P}(S)$ as the set of probability measures on $S$. For $x \in \mathbb{R}^d$, $\| x \|$ refers to the $L^2$ norm of $x$ while $B_2(x, \epsilon)$ denotes the closed $L^2$ ball centered at $x$ with width $\epsilon$. For a collection of random variables $(Z_1, \cdots, Z_t)$, $\sigma(Z_1, \cdots, Z_t)$ refers to the sigma-field generated by $Z_1, \cdots, Z_t$. For a random variable $Z$ and a probability distribution $p$, we write $Z \sim p$ if $Z$ is distributed according to $p$.
\section{Lower bounds}
\label{sec-lower-bound}
Unless otherwise stated, we assume throughout this section that $\ell$ can be written in the form \eqref{eq-def-general-loss}. In particular, $\ell(z, \cdot)$ is continuous for any $z \in \mathcal{Z}$. We build on a powerful result rooted in von Neumann's minimax theorem that enables the derivation of tight lower and upper bounds on $R_T\stdgame$ by recasting the value of the game in a backward order.

\begin{theorem} From \cite{abernethy2009stochastic}
\label{lemma-minimax-reformulation}
\begin{equation*}
	R_T\stdgame = \sup\limits_{p} \mathbb{E} [ \; \sum_{t=1}^T  \inf_{f_t \in \mathcal{F}} \mathbb{E}[ \ell(Z_t, f_t) | Z_1, \cdots, Z_{t-1} ] - \inf\limits_{f \in \mathcal{F}} \sum_{t=1}^T \ell(Z_t, f) \; ],
\end{equation*}
where the supremum is taken over the distribution $p$ of the random variable $(Z_1, \cdots, Z_T)$ in $\mathcal{Z}^T$. 
\end{theorem}

Any choice for $p$ yields a lower bound on $R_T\stdgame$. The following result identifies a canonical choice for $p$ that leads to $\omegasqrt$ lower bounds on regret.

\begin{lemma} Adapted from \cite{abernethy2009stochastic} \\
\label{lemma-tool-to-prove-LB}
If we can find a distribution $p$ on $\mathcal{Z}$ and two points $f_1$ and $f_2$ in $\argminf \mathbb{E}[\ell(Z, f)]$ such that $\ell(Z, f_1) \neq \ell(Z, f_2)$ with positive probability for $Z \sim p$, then $R_T\stdgame = \omegasqrt$. 
\end{lemma}
A distribution $p$ satisfying the requirements of Lemma \ref{lemma-tool-to-prove-LB} can be viewed as an equalizing strategy for the opponent. This concept, formalized in \cite{Rakhlin2011}, roughly refers to randomized strategies played by the opponent that cause the player's decisions to be completely irrelevant from a regret standpoint. These strategies are intrinsically hard to play against and often lead to tight lower bounds. To gain some intuition about this result, suppose that that the opponent generates an independent copy of $Z$ at each round $t$, which we denote by $Z_t$. In the adversarial setting considered in this paper, the player is aware of the opponent's strategy but does not get to see the realization of $Z_t$ before committing to a decision. For this reason, at any round, $f_1$ and $f_2$ are optimal moves that are completely equivalent from the player's perspective. However, in hindsight, i.e. once all the realizations of the $Z_t$'s have been revealed, $f_1$ and $f_2$ are typically not equivalent because $\ell(Z_t, f_1) \neq \ell(Z_t, f_2)$ with positive probability and one of these two moves will turn out to be 
\begin{equation*}
	\max(0, \sum_{t=1}^T \ell(Z_t, f_1) - \ell(Z_t, f_2))
\end{equation*}
suboptimal which, in expectations, is of order $\omegasqrt$ by the central limit theorem. Given the conditions imposed on $p$, it is convenient to work with the following equivalence relation.

\begin{definition}
\label{definition-equivalenceClass}
We define the equivalence relation $\sim_{\ell}$ on $\mathcal{F}$ by $f_a \sim_{\ell} f_b$ for $f_a, f_b \in \mathcal{F}$ if and only if $\ell(z, f_a) = \ell(z, f_b)$ for all $z \in \mathcal{Z}$.
\end{definition}

In what follows, we show, using sensitivity analysis for linear programming, that we can systematically, with the only exception of trivial games defined below, construct a distribution $p$ with support $\mathcal{Z}$ such that there are at least two equivalent classes in $\argminf \mathbb{E}[\ell(Z, f)]$ whenever $\mathcal{F}$ is a polyhedron, for $Z \sim p$.

\begin{definition}
\label{definition-GameIsTrivial}
The game $\stdgame$ is said to be trivial if and only if
\begin{equation*}
\exists f^* \in \mathcal{F} \; \text{such that} \; \forall z \in \mathcal{Z}, \; \ell(z, f^*) \leq \min_{f \in \mathcal{F}} \ell(z, f). 
\end{equation*}
\end{definition}
A simple example of a trivial game is $(\ell(z, f) = z f, [0, 1], [0, 1])$, where $\ell(z, f) \geq 0$ $\forall f \in [0,1]$ and $\forall z \in [0, 1]$, with $\ell(z, f) = 0$ if $f = 0$ irrespective of $z$. If the game is trivial, the player will always play $f^*$ irrespective of the time horizon and of the opponent's strategy observed so far to obtain zero regret. As it turns out, this uniquely identifies trivial games as we establish in Lemma \ref{lemma-ZeroRegretImpliesTrivial}. 

\begin{lemma}
\label{lemma-ZeroRegretImpliesTrivial}
For any $T \in \mathbb{N}$, $R_T\stdgame \geq 0$. Moreover in any of the following cases:
\begin{enumerate}
	\item $\mathcal{Z}$ has finite cardinality,
	\item $\ell(\cdot, f)$ is continuous for any choice of $f \in \mathcal{F}$,
\end{enumerate}
$R_T\stdgame = 0$ if and only if the game is trivial.
\end{lemma}

The following result shows that, in most cases of interest, we can drastically restrict the power of the opponent while still preserving the nature of the game. This enables us to focus on the case where $\mathcal{Z}$ is finite.
\begin{lemma}
\label{lemma-non-trivial-imply-finite-non-trivial}
Suppose that $\ell(\cdot, f)$ is continuous for any choice of $f \in \mathcal{F}$. If the game $\stdgame$ is not trivial, there exists a finite subset $\tilde{\mathcal{Z}} \subseteq  \mathcal{Z}$ such that the game $(\ell, \tilde{\mathcal{Z}}, \mathcal{F})$ is not trivial.
\end{lemma}

We are now ready to present the main results of this section. To the best of our knowledge, these results constitute the first systematic $\omegasqrt$ lower bounds on regret obtained for a large class of piecewise linear loss functions.
\begin{theorem}
\label{lemma-OmegaSqrtPiecewiseLinear}
Suppose that $\mathcal{F}$ is a polyhedron. In any of the following cases:
\begin{enumerate}
	\item $\mathcal{Z}$ has finite cardinality,
	\item $\ell(\cdot, f)$ is continuous for any choice of $f \in \mathcal{F}$,
\end{enumerate}
either the game is trivial or $R_T\stdgame = \omegasqrt$.
\end{theorem}

An immediate consequence of Theorem \ref{lemma-OmegaSqrtPiecewiseLinear} for linear games is the following:
\begin{theorem}
\label{lemma-OmegaSqrtLinearPolyhedron}
Suppose that $\mathcal{F}$ is a polyhedron and that $\ell(z, f) = z^\tr f$. Then, either the game is trivial or $R_T\stdgame = \omegasqrt$.
\end{theorem}

The proofs rely on Lemma \ref{lemma-tool-to-prove-LB} which is based on Theorem \ref{lemma-minimax-reformulation} and may, as a result, seem rather obscure. We stress that these lower bounds are derived by means of an equalizing strategy. We present this more intuitive view in the Appendix by exhibiting an equalizing strategy in the online linear optimization setting when $\interior( \conv(\mathcal{Z}) ) \neq \emptyset$. \\
Note that Theorems \ref{lemma-OmegaSqrtPiecewiseLinear} and \ref{lemma-OmegaSqrtLinearPolyhedron} imply $\omegasqrt$ regret for a number of repeated Stackelberg games and online linear optimization problems as discussed in Section \ref{sec-applications}. Furthermore, we stress that Theorem \ref{lemma-OmegaSqrtPiecewiseLinear} can also be used when $\mathcal{F}$ is not a polyhedron but this typically requires a preliminary step which boils down to restricting the opponent's decision set. For instance, the following well-known result is almost a direct consequence of Theorem \ref{lemma-OmegaSqrtLinearPolyhedron}.

\begin{lemma}
\label{lemma-ZeroInInterior}
Suppose that $\ell(z, f) = z^\tr f$, that $0 \in \interior( \conv( \mathcal{Z} ) )$, and that $\mathcal{F}$ contains at least two elements. Then $R_T\stdgame = \omegasqrt$.
\end{lemma}
Note that Lemma \ref{lemma-ZeroInInterior} is consistent with Theorem \ref{lemma-OmegaSqrtLinearPolyhedron} as the game $(\ell(z, f) = z^\tr f, \mathcal{Z}, \mathcal{F})$ is non-trivial if $0 \in \interior( \conv( \mathcal{Z} ) )$ as soon as $\mathcal{F}$ contains at least two elements. Indeed:
\begin{equation*}
	\ell(\epsilon \frac{f_2 - f_1}{\| f_2 - f_1\| }, f_2) > \ell(\epsilon \frac{f_2 - f_1}{\| f_2 - f_1\| }, f_1) \; \text{and} \; \ell(\epsilon \frac{f_1 - f_2}{\| f_2 - f_1\| }, f_1) > \ell(\epsilon \frac{f_1 - f_2}{\| f_2 - f_1\| }, f_2), 
\end{equation*}
for a small enough $\epsilon >0$ and any pair $f_1 \neq f_2 \in \mathcal{F}$. When $0 \in \interior( \mathcal{Z} )$, the opponent has some freedom to play, at each time period, a random vector with expected value zero, making every strategy available to the player equally bad. In other words, any i.i.d. zero-mean distribution is an equalizing strategy for the opponent in this case.
\\
A preliminary step is also required to derive $\omegasqrt$ lower bounds on regret for prediction problems with side information where $\mathcal{F}$ is typically not a polyhedron. We sketch this simple argument for the canonical classification problem with the hinge loss, i.e. the game :
\begin{equation*}
	(\ell((x, y), f) = \max(0, 1 - y x^\tr f ),  \mathcal{Z} = B_2(0, 1) \times \{-1, 1\}, \mathcal{F} = B_2(0, 1)),
\end{equation*}
but the method readily extends to any of the prediction problems described in Section \ref{sec-applications}. The idea is to restrict the opponent's decision set by taking a fixed vector $x$ of norm 1 and impose that, at any round $t$, the opponent's move be $(x, y_t)$ for $y_t \in \{-1, 1 \}$. Since $\ell((x, y), f)$ only depends on $f$ through the scalar product between $f$ and $x$, the player's decision set can be equivalently described by this value, which lies in $[-1, 1]$. Formally, we define a new loss function $\tilde{\ell}(y, f) = \max(0, 1 - y f)$ with $\tilde{\mathcal{Z}} = \{-1, 1\}$ and $\tilde{\mathcal{F}} = [-1, 1]$ and we have:
\begin{equation*}
	R_T(\ell, \mathcal{Z}, \mathcal{F}) \geq R_T(\tilde{\ell}, \tilde{\mathcal{Z}}, \tilde{\mathcal{F}}).
\end{equation*}
Observe now that the game $(\tilde{\ell}, \tilde{\mathcal{Z}}, \tilde{\mathcal{F}})$ is not trivial, that $\mathcal{Z}$ is discrete, and that:
\begin{equation*}
		\tilde{\ell}(y, f) = \max_{\alpha \in \{0, 1\}} (\alpha, -y \alpha)^\tr (1, f).
\end{equation*}
We conclude with Theorem \ref{lemma-OmegaSqrtPiecewiseLinear} that $R_T(\tilde{\ell}, \tilde{\mathcal{Z}}, \tilde{\mathcal{F}}) = \omegasqrt$ which implies that $R_T(\ell, \mathcal{Z}, \mathcal{F}) = \omegasqrt$.

\paragraph{Remark about Lemma \ref{lemma-tool-to-prove-LB}}
We point out that, in general, it is not possible to weaken the assumptions of Lemma \ref{lemma-tool-to-prove-LB} (which, in fact, applies to a much more general class of loss functions than the one given by \eqref{eq-def-general-loss}). In particular, finding $z \in \mathcal{Z}$ such that there are two equivalence classes $f_1$ and $f_2$ in $\argminf \ell(z, f)$ does not guarantee that $R_T\stdgame = \omegasqrt$ as we illustrate with a counterexample. This is because the result of Lemma \ref{lemma-tool-to-prove-LB} is intrinsically tied to the central limit theorem. Consider the following (non-trivial) online linear regression game:
\begin{equation*}
	(\ell(z, f) = (z^\tr f)^2, \mathcal{Z} = B_2(z^*, 1), \mathcal{F} = [f_1, f_2]),
\end{equation*}
where $f_1 = (1, 0, \cdots, 0)$, $f_2 = (0, 1, 0, \cdots)$, and $z^* = (1, 1, 0, \cdots, 0)$. Observe that $\forall z \in \mathcal{Z}, \forall f \in \mathcal{F}, z^\tr f \geq 0$. Hence $\argminf \ell(z, f) = \argminf f^\tr z$. Furthermore, $\argminf f^\tr z^* = [f_1, f_2]$ but $f_1$ and $f_2$ are clearly not in the same equivalence class. Yet, even though $\ell$ is not strongly convex in $f$, there exists an algorithm achieving $O( \log(T) )$ regret, see \cite{vovk1998competitive}. 

\section{Upper bounds}
\label{sec-upper-bound}
Looking at Theorem \ref{lemma-OmegaSqrtPiecewiseLinear}, Theorem \ref{lemma-OmegaSqrtLinearPolyhedron}, and Lemma \ref{lemma-ZeroInInterior}, it is tempting to conclude that the existence of pieces where $\ell$ is linear in its second argument dooms us to $\Omega(\sqrt{T})$ regret bounds. We argue that the growth rate of $R_T\stdgame$ is determined by a more involved interplay between $\ell$, $\mathcal{Z}$, and $\mathcal{F}$ so that this assertion requires further examination. In fact, we show that $O(\log(T))$ regret bounds are even possible in online linear optimization. The fundamental reason is that the curvature of the boundary of $\mathcal{F}$ can make up for the lack of thereof of $\ell$. Curvature is key to enforce stability of the player's strategy with respect to perturbations in the opponent's moves. Sometimes, when the predictions are stable, e.g. when $\ell$ is the square loss $\ell(z, f) = \norm{z - f}^2$, a very simple algorithm, known as Follow-The-Leader, yields $O(\log(T))$ regret.

\begin{definition} From \cite{kalai2005efficient} \\
\label{def-FTL}
The Follow-The-Leader (FTL) strategy consists in playing:
\begin{equation*}
			f_t \in \argmin_{f \in \mathcal{F}} \frac{1}{t-1} \sum_{\tau = 1}^{t-1} \ell(z_\tau, f).
\end{equation*}
\end{definition}
It is well-known that FTL fails to yield sublinear regret for online linear optimization in general. However, when $\mathcal{F}$ has a strongly curved boundary and $0 \notin \conv(\mathcal{Z})$, the FTL strategy becomes stable, leading to $O(\log(T))$ regret as we next show using sensitivity analysis.

\begin{theorem}
\label{lemma-ftl-yields-logT}
Suppose that (i) $\ell$ is linear, (ii) $\mathcal{F} = \{f \in \mathbb{R}^{d_f} \; | \; F(f) \leq 0 \}$ for $F$ a strongly convex function with respect to the $L^2$ norm, and (iii) $0 \notin \conv(\mathcal{Z})$. Then, FTL yields $O(\log(T))$ regret.
\end{theorem}
As an example of application of Theorem \ref{lemma-ftl-yields-logT}, consider a repeated network game where the player picks the arc costs in a $L^2$ ball, the opponent picks a path, and the loss incurred to the opponent is the sum of the arc costs along the path. In this setting, FTL yields $O(\log(T))$ regret even though the game is not trivial. Theorem \ref{lemma-ftl-yields-logT} also has some implications for non-linear convex loss functions when the boundary of $\mathcal{F}$ is curved and $0$ is not in the convex hull of the set of subgradients of $\ell$ with respect to the player's moves. Indeed suppose that, at any time period $t$, the player follows the FTL strategy as if the loss function were linear and the past moves of the opponents were $y_1, \cdots, y_{t-1}$, i.e.: 
\begin{equation*}
			f_t \in \argmin_{f \in \mathcal{F}} \frac{1}{t-1} \sum_{\tau = 1}^{t-1} y_\tau^\tr f,
\end{equation*}
where, for any $\tau = 1, \cdots, t-1$, $y_\tau$ is a subgradient of $\ell(z_t, \cdot)$ at $f_t$. Then, for any sequence of moves $(z_1, \cdots, z_T)$, we have:
\begin{align*}
	r_T((z_t)_{t=1, \cdots, T}, (f_t)_{t=1, \cdots, T}) 
		& \leq \sum_{t=1}^T y_t^\tr f_t - \inf\limits_{f \in \mathcal{F}} \sum_{t=1}^T y_t^\tr f \\
		& = O(\log(T)).
\end{align*}
It is, however, a priori unclear whether $\log(T)$ is the optimal growth rate for games satisfying the assumptions of Theorem \ref{lemma-ftl-yields-logT}. Quite surprisingly, i.i.d. opponents appear to be particularly weak for this kind of games, incurring at most a $O(1)$ regret lower bound as shown in the following lemma. This is in stark contrast with the situations of Section \ref{sec-lower-bound} where the (tight) $\omegasqrt$ lower bounds are always derived through i.i.d. opponents. 
\begin{lemma}
\label{lemma-NotEnoughIID}
Consider the game $(\ell(z, f) = z^\tr f, \mathcal{Z}, \mathcal{F})$ with $0 \notin \conv(\mathcal{Z})$ and $\mathcal{F} = B_2(0, 1)$. Any lower bound derived from Theorem \ref{lemma-minimax-reformulation} with i.i.d. random variables $Z_1, \cdots, Z_T \sim p$ is $O(1)$ for any choice of $p \in \mathcal{P}( \mathcal{Z} )$.
\end{lemma}
\citet{abernethy2009stochastic} remark that restricting the study to i.i.d. sequences is in general not enough to get tight bounds for non-linear losses such as $\ell(z, f) = \norm{z - f}^2$. It turns out that this is also true for linear losses when $\mathcal{F}$ is strongly curved and $0 \notin \conv(\mathcal{Z})$ as the value of the game is in fact $\Theta(\log(T))$.
\begin{theorem}
\label{lemma-NonIIDLowerBound}
When $\ell(z, f) = z^\tr f$, $0 \notin \conv(\mathcal{Z})$, and $\mathcal{F} = B_2(0, 1)$, the value of the game is:
$$
	R_T\stdgame = \Theta(\log(T)).
$$
\end{theorem}
So far, we have studied two scenarios that are diametrically opposed in terms of the curvature of the decision sets with polyhedra on one side in Section \ref{sec-lower-bound}, with $\Theta(\sqrt{T})$ regret, and euclidean balls, with $\Theta(\log(T))$ regret, on the other side of the spectrum. Interestingly, bridging this gap leads to the rise of intermediate learning rates that can be quantified through the modulus of convexity of $\mathcal{F}$. Specifically, consider any norm $\norm{ \; }_{\mathcal{F}}$. The modulus of convexity of the associated unit ball is defined as:
$$
	\delta_{\mathcal{F}} \vcentcolon \epsilon \rightarrow \inf\limits_{ \substack{ \norm{f}_\mathcal{F}, \norm{\tilde{f}}_\mathcal{F} \leq 1 \\ \norm{f - \tilde{f}}_{\mathcal{F}} \geq \epsilon} } 1 - \norm{\frac{f + \tilde{f}}{2}}_\mathcal{F}.
$$
The norm $\norm{ \;}_\mathcal{F}$ is said to be uniformly convex if its characteristic of convexity is equal to $0$, i.e.:
$$
	\sup \{ \epsilon \geq 0 \; | \; \delta_\mathcal{F}(\epsilon) = 0 \} = 0.
$$
\citet{pisier1975martingales} show that if $\norm{ \;}_\mathcal{F}$ is uniformly convex, there must exist $q \geq 2$ and $c > 0$ such that $\delta_\mathcal{F}(\epsilon) \geq  c \epsilon^q$ for all $\epsilon \in [0, 2]$, in which case we say that $\norm{ \;}_\mathcal{F}$ is $q-$uniformly convex. This parameter quantifies how curved $\norm{\;}_\mathcal{F}$ balls are and determines the growth rate of the value of the game when $\mathcal{F}$ is a $\norm{\;}_\mathcal{F}$ ball and $0 \notin \conv(\mathcal{Z})$.
\begin{lemma}
\label{lemma-UniConvex-UpperBound}
Consider the game $(\ell(z, f) = z^\tr f, \mathcal{Z}, \mathcal{F})$ with $0 \notin \conv(\mathcal{Z})$ and $\mathcal{F} = \{f \; | \; \norm{f}_{\mathcal{F}} \leq C \}$, where $\norm{ \;}_\mathcal{F}$ is a $q$-uniformly convex norm and $C \geq 0$. Then, FTL yields regret $O(\log(T))$ if $q=2$ and regret $O( T^{\frac{q-2}{q-1}} )$ if $q \in (2, 3]$.
%$$
%	 \begin{array}{lcl}
%						 O(\log(T)) & \mbox{if} & q=2 \\
%						 O( T^{\frac{q-2}{q-1}} ) & \mbox{if} & q \in (2, 3]
%				  \end{array} 
%	.
%$$
\end{lemma}

\paragraph{Situations where $0$ lies on the boundary of $\mathcal{Z}$ when $\ell$ is linear} Observe that this situation is not covered by Lemma \ref{lemma-ZeroInInterior}, Theorem \ref{lemma-ftl-yields-logT}, Theorem \ref{lemma-NonIIDLowerBound}, nor Lemma \ref{lemma-UniConvex-UpperBound}. We stress that zero-mean i.i.d. opponents are not helpful to derive $\omegasqrt$ regret when $0$ lies exactly on the boundary of $\mathcal{Z}$ (in fact, in the relative interior of an edge of $\mathcal{Z}$), as we illustrate with an example. Therefore, the growth rate of $R_T\stdgame$ remains unknown in this setup. Define $\mathcal{Z} = \conv(z_1, z_2, z_3, z_4)$ with $z_1 = (-1, 1, 0, 0)$, $z_2 = (1, -1, 0, 0)$, $z_3 = (0, 0, 0, 1)$, and $z_4 = (0, 0, 1, 0)$. Also define $\mathcal{F} = [f^*, f^{**}]$ with $f^* = (0, 0, 0, 0)$ and $f^{**} = (1, 1, -1, 1)$. Observe that the game $(\ell(z, f) = z^\tr f, \mathcal{Z}, \mathcal{F})$ is not trivial because $\argminf f \cdot z_3 = \{ f^{*}\}$ while $\argminf f \cdot z_4 = \{ f^{**} \}$. For any zero-mean i.i.d. opponent $Z_1, \cdots, Z_T$, the only possibility is to have $Z_t \in [z_1, z_2]$. We get, irrespective of the player's strategy:
\begin{equation*}
\mathbb{E}[r_T((Z_t)_{t=1, \cdots, T}, (f_t)_{t=1, \cdots, T})] = - \mathbb{E}[ \inf\limits_{f \in \mathcal{F}} f \cdot \sum_{t=1}^T Z_t].
\end{equation*}
We have $f^* \in \argminf f \cdot z$ for any $z \in [z_1, z_2]$. Hence:
\begin{equation*}
	\mathbb{E}[r_T((Z_t)_{t=1, \cdots, T}, (f_t)_{t=1, \cdots, T})] = - \mathbb{E}[ f^* \cdot \sum_{t=1}^T Z_t],
\end{equation*}
which finally yields $\mathbb{E}[r_T((Z_t)_{t=1, \cdots, T}, (f_t)_{t=1, \cdots, T})] = 0$. 

\section*{Acknowledgments}
The authors would like to thank Alexander Rakhlin for his valuable input, and in particular, for bringing to our attention the possibility of having $o(\sqrt{T})$ bounds on regret in the linear setting.

\bibliographystyle{plainnat}

\bibliography{biblioOnlineOptimization}

\section{Appendix: proofs}
\subsection{Proof of Theorem \ref{lemma-minimax-reformulation}}
The assumptions of Theorem 1 in \cite{abernethy2009stochastic} are satisfied for the game $\stdgame$ using Assumption \ref{assumption-sets-all} and the fact that any loss function $\ell$ of the form \eqref{eq-def-general-loss} is such that $\ell(z, \cdot)$ is continuous for any $z \in \mathcal{Z}$.

\subsection{Proof of Lemma \ref{lemma-LinearConvexGameIsEquivalent}}

	The proof follows from the repeated use of the Von Neumann's minimax theorem developed in \cite{abernethy2009stochastic}. To simplify the presentation, we prove the result when $T=2$ but the general proof follows the same principle. We have:
	$$
		R_2 = \inf\limits_{f_1 \in \mathcal{F}} \sup\limits_{z_1 \in \mathcal{Z}} \inf\limits_{f_{2} \in \mathcal{F}} \sup\limits_{z_{2} \in \mathcal{Z}} [ \; \sum_{t=1}^2 \ell(z_t, f_t) - \inf\limits_{f \in \mathcal{F}} \sum_{t=1}^2 \ell(z_t, f) \; ].
	$$
	Consider fixed vectors $f_1, f_2 \in \mathcal{F}$ and $z_1 \in \mathcal{Z}$ and define the function $M(z_2) = \sum_{t=1}^2 \ell(z_t, f_t) - \inf_{f \in \mathcal{F}} \sum_{t=1}^2 \ell(z_t, f)$. Observe that $M$ is convex. Indeed, $z_2 \rightarrow \ell(z_1, f_1) + \ell(z_2, f_2)$ is affine and $z_2 \rightarrow \inf_{f \in \mathcal{F}} \sum_{t=1}^2 \ell(z_t, f)$ is concave as the infimum of affine functions. Therefore: 
	$$
	\sup_{z_2 \in \mathcal{Z}} M(z_2) = \sup_{z_2 \in \conv( \mathcal{Z} )} M(z_2).
	$$
	We obtain:
	$$
		R_2 = \inf\limits_{f_1 \in \mathcal{F}} \sup\limits_{z_1 \in \mathcal{Z}} \inf\limits_{f_{2} \in \mathcal{F}} \sup\limits_{z_{2} \in \conv( \mathcal{Z} )} [ \; \sum_{t=1}^2 \ell(z_t, f_t) - \inf\limits_{f \in \mathcal{F}} \sum_{t=1}^2 \ell(z_t, f) \; ].
	$$
	By randomizing the choice of $z_2$, we can use Von Neumann's minimax theorem to derive:
	$$
		R_2 = \inf\limits_{f_1 \in \mathcal{F}} \sup\limits_{z_1 \in \mathcal{Z}} \; \{ \; \ell(z_1, f_1) + \sup\limits_{p_2 \in \mathcal{P}( \conv( \mathcal{Z} ) )} \; \{ \; \inf\limits_{f_2 \in \mathcal{F} } \mathbb{E}_{z \sim p_2} \ell(z, f_2) - \mathbb{E}_{z_2 \sim p_2} \inf\limits_{f \in \mathcal{F}} \sum_{t=1}^2 \ell(z_t, f) \; \} \; \}.
	$$
	For a fixed $f_1 \in \mathcal{F}$, define:
	$$
	A(z_1) = \; \ell(z_1, f_1) + \sup\limits_{p_2 \in \mathcal{P}( \conv( \mathcal{Z} ) )} \; \{ \; \inf\limits_{f_2 \in \mathcal{F} } \mathbb{E}_{z \sim p_2} \ell(z, f_2) - \mathbb{E}_{z_2 \sim p_2} \inf\limits_{f \in \mathcal{F}} \sum_{t=1}^2 \ell(z_t, f) \; \}.
	$$ 
	Observe that, for a fixed $p_2 \in \mathcal{P}( \conv( \mathcal{Z} ) )$, the function: 
	$$
	z_1 \rightarrow \inf_{f_2 \in \mathcal{F} } \mathbb{E}_{z \sim p_2} \ell(z, f_2) - \mathbb{E}_{z_2 \sim p_2} \inf_{f \in \mathcal{F}} \sum_{t=1}^2 \ell(z_t, f)
	$$ 
	is convex as the difference between a constant and the expected value of the infimum of affine functions. Since the supremum of convex functions is convex, $A$ is convex and $\sup_{z_1 \in \mathcal{Z}} A(z_1) = \sup_{z_1 \in \conv(\mathcal{Z})} A(z_1)$. We derive:
	$$
		R_2 = \inf\limits_{f_1 \in \mathcal{F}} \sup\limits_{z_1 \in \conv( \mathcal{Z} )} [ \; \ell(z_1, f_1) + \sup\limits_{p_2 \in \mathcal{P}( \conv( \mathcal{Z} ) )} \; \{ \; \inf\limits_{f_2 \in \mathcal{F} } \mathbb{E}_{z	 \sim p_2} \ell(z, f_2) - \mathbb{E}_{z_2 \sim p_2} \inf\limits_{f \in \mathcal{F}} \sum_{t=1}^2 \ell(z_t, f) \} ].
	$$
	To conclude, we unwind the first step, i.e. we use the minimax theorem in reverse order. This yields:
	$$
		R_2 = \inf\limits_{f_1 \in \mathcal{F}} \sup\limits_{z_1 \in \conv( \mathcal{Z} )} \inf\limits_{f_{2} \in \mathcal{F}} \sup\limits_{z_{2} \in \conv( \mathcal{Z} )} [  \sum_{t=1}^2 \ell(z_t, f_t) - \inf\limits_{f \in \mathcal{F}} \sum_{t=1}^2 \ell(z_t, f) \; ],
	$$	
	i.e. $R_2(\ell(z, f) = z^\tr f, \mathcal{Z}, \mathcal{F}) = R_2(\ell(z, f) = z^\tr f, \conv( \mathcal{Z} ), \mathcal{F})$. Moreover, $\mathcal{Z}$ is a compact set which implies that $\conv( \mathcal{Z} )$ is also a compact set by a standard topological argument. As a result, the game $(\ell(z, f) = z^\tr f, \conv( \mathcal{Z} ), \mathcal{F})$ also satisfies Assumption \ref{assumption-sets-all}.

\subsection{Proof of Lemma \ref{lemma-tool-to-prove-LB}}
We follow the analysis of Theorem 19 of \cite{abernethy2009stochastic}. Using Theorem \ref{lemma-minimax-reformulation} with $p$ taken as the distribution of i.i.d. copies of $Z$, we get the lower bound:
\begin{align*}
	R_T
		& \geq T \inf_{f \in \mathcal{F}} \mathbb{E}[ \ell(Z_t, f) ] - \mathbb{E}[ \inf\limits_{f \in \mathcal{F}} \sum_{t=1}^T \ell(Z_t, f) ] \\
		& \geq T \sup_{f \in \{f_1, f_2\}} \mathbb{E}[ \ell(Z_t, f) ] - \mathbb{E}[ \inf\limits_{f \in \{f_1, f_2\}} \sum_{t=1}^T \ell(Z_t, f) ] \\
		& \geq \mathbb{E}[ \max\{\sum_{t=1}^T \mathbb{E}[ \ell(Z_t, f_1)] - \ell(Z_t, f_1), \sum_{t=1}^T \mathbb{E}[ \ell(Z_t, f_2)] - \ell(Z_t, f_2) \} ] \\
		& \geq \mathbb{E}[ \max\{0, \sum_{t=1}^T \ell(Z_t, f_1)- \ell(Z_t, f_2) \} ],
\end{align*}
where we use the fact that $\inf_{f \in \mathcal{F}} \mathbb{E}[ \ell(Z_t, f) ] = \mathbb{E}[ \ell(Z_t, f_1)] = \mathbb{E}[ \ell(Z_t, f_2)]$. Since $\ell(Z, f_2) \neq \ell(Z, f_1)$ with positive probability, the random variables $(\ell(Z_t, f_1)- \ell(Z_t, f_2))_{t=1, \cdots, T}$ are i.i.d. with zero mean and positive variance and we can conclude with the central limit theorem since $\ell$ is bounded.

\subsection{Proof of Lemma \ref{lemma-ZeroRegretImpliesTrivial}}

The fact that $R_T \geq 0$ is proved in Lemma 3 of \cite{abernethy2009stochastic} and follows from Theorem \ref{lemma-minimax-reformulation} by taking the $Z_t$'s to be deterministic and all equal to any $z \in \mathcal{Z}$. Clearly, if the game is trivial then $R_T = 0$ because this value is attained for $f_1, \cdots, f_T = f^*$ irrespective of the decisions made by the opponent. Conversely, suppose $R_T = 0$. Consider $p$ to be the product of $T$ uniform distributions on $\mathcal{Z}$. Then, using again Theorem \ref{lemma-minimax-reformulation}:
$$
0 \geq \mathbb{E} [ \; \sum_{t=1}^T  \inf_{f_t \in \mathcal{F}} \mathbb{E}[ \ell(Z_t, f_t)] - \inf\limits_{f \in \mathcal{F}} \sum_{t=1}^T \ell(Z_t, f) \; ],
$$ 
as $Z_1, \cdots, Z_T$ are independent random variables. Since they are also identically distributed, we obtain:
$$
0 \geq T\cdot \inf_{f \in \mathcal{F}} \mathbb{E}[ \ell(Z, f)] - \mathbb{E}[ \inf\limits_{f \in \mathcal{F}} \sum_{t=1}^T \ell(Z_t, f)].
$$
Yet $\mathbb{E} [ \inf\limits_{f \in \mathcal{F}} \sum_{t=1}^T \ell(Z_t, f) ] \leq \inf\limits_{f \in \mathcal{F}} \mathbb{E}[ \sum_{t=1}^T \ell(Z_t, f) ] = T \cdot \inf\limits_{f \in \mathcal{F}} \mathbb{E}[ \ell(Z, f)]$ and we derive:
 $$
	T\cdot \inf_{f \in \mathcal{F}} \mathbb{E}[ \ell(Z, f)] - \mathbb{E}[ \inf\limits_{f \in \mathcal{F}} \sum_{t=1}^T \ell(Z_t, f)] = 0.
$$
Since $\ell$ is bounded, $\mathcal{Z}$ is compact, and $\ell(z, \cdot)$ is continuous for any $z \in \mathcal{Z}$, $f \rightarrow \mathbb{E}[\ell(Z, f)]$ is continuous by dominated convergence so we can take $f^* \in \argminf \mathbb{E}[ \ell(Z, f)]$ ($\mathcal{F}$ is compact). We obtain: 
$$
\mathbb{E} [ \sum_{t=1}^T \ell(Z_t, f^*) - \inf\limits_{f \in \mathcal{F}} \sum_{t=1}^T \ell(Z_t, f) ] = 0.
$$
As $\sum_{t=1}^T \ell(Z_t, f^*) - \inf\limits_{f \in \mathcal{F}} \sum_{t=1}^T \ell(Z_t, f) \geq 0$, we derive that:
$$
(z_1, \cdots, z_T) \rightarrow \sum_{t=1}^T \ell(z_t, f^*) - \inf\limits_{f \in \mathcal{F}} \sum_{t=1}^T \ell(z_t, f) = 0
$$ 
almost everywhere on $\mathcal{Z}^T$. If $\mathcal{Z}$ is discrete, this implies equality on $\mathcal{Z}^T$, which in particular implies $\ell(z, f^*) = \inf\limits_{f \in \mathcal{F}} \ell(z, f)$ for all $z \in \mathcal{Z}$ and we are done. If, on the other hand, $\ell(\cdot, f)$ is continuous for all $f \in \mathcal{F}$, we have:
$$
	\sum_{t=1}^T \ell(z_t, f^*) \leq \sum_{t=1}^T \ell(z_t, f), \; \forall f \in \mathcal{F},  \forall (z_1, \cdots, z_T) \in \tilde{\mathcal{Z}},
$$
for $\tilde{\mathcal{Z}}$ a subset of $\mathcal{Z}$ with Lebesgue measure equal to that of $\mathcal{Z}$. Since a non-empty open set  cannot have Lebesgue measure $0$, $\tilde{\mathcal{Z}}$ is dense in $\mathcal{Z}$ and by taking limits in the above inequality for each $f \in \mathcal{F}$ separately, we conclude that: 
$$
	\sum_{t=1}^T \ell(z_t, f^*) \leq \sum_{t=1}^T \ell(z_t, f), \; \forall f \in \mathcal{F}, \forall (z_1, \cdots, z_T) \in \mathcal{Z},
$$
which in particular implies that $\ell(z, f^*) = \inf\limits_{f \in \mathcal{F}} \ell(z, f)$ for all $z \in \mathcal{Z}$ and the game is trivial.

\subsection{Proof of Lemma \ref{lemma-non-trivial-imply-finite-non-trivial}}

Suppose by contradiction that we cannot find such a finite subset. Since $\mathcal{Z}$ is compact, it is also separable thus it contains a countable dense subset $\{ z_n \; | \; n \in \mathbb{N} \}$. By assumption, the game $(\ell, \{ z_k \; | \; k \leq n \}, \mathcal{F})$ must be trivial for any $n$, i.e. there exists $f_n \in \mathcal{F}$ such that:
$$
	\ell(z_k, f_n) \leq \min_{f \in \mathcal{F}} \ell(z, f), \; \forall k \leq n.
$$
Since $\mathcal{F}$ is compact, we can find a subsequence of $(f_n)_{n \in \mathbb{N}}$ such that $f_n \rightarrow f^* \in \mathcal{F}$. Without loss of generality, we continue to refer to this sequence as $(f_n)_{n \in \mathbb{N}}$. Taking the limit $n \rightarrow \infty$ in the above inequality for any fixed $k \in \mathbb{N}$ yields:
$$
	\ell(z_k, f^*) \leq \ell(z_k, f), \; \forall f \in \mathcal{F}, \forall k \in \mathbb{N}.
$$
Consider a fixed $f \in \mathcal{F}$, since $\{ z_n \; | \; n \in \mathbb{N} \}$ is dense in $\mathcal{Z}$ and since $\ell(\cdot, f^*)$ and $\ell(\cdot, f)$ are continuous, we get:
$$
	\ell(z, f^*) \leq \ell(z, f), \; \forall f \in \mathcal{F}, \forall z \in \mathcal{Z},
$$
which shows that $\stdgame$ is trivial, a contradiction.

\subsection{Proof of Theorem \ref{lemma-OmegaSqrtPiecewiseLinear}}

Without loss of generality we can assume that the game is not trivial and that $\mathcal{X}(z)$ is finite for any $z \in \mathcal{Z}$ since otherwise if $\mathcal{X}(z)$ is a polyhedron, the maximum in \eqref{eq-def-general-loss} must be attained at an extreme point of $\mathcal{X}(z)$ ($\ell$ is bounded by Assumption \ref{assumption-sets-all}) and there are finitely many such points for any $z$. Moreover, we can also assume that $\mathcal{Z}$ is discrete by Lemma \ref{lemma-non-trivial-imply-finite-non-trivial} since, borrowing the notations of Lemma \ref{lemma-non-trivial-imply-finite-non-trivial}, we have:
$$
	R_T(\ell, \mathcal{Z}, \mathcal{F}) \geq R_T(\ell, \tilde{\mathcal{Z}}, \mathcal{F}).
$$
Write $\mathcal{Z} = \{ z_n \; | \; n \leq N \}$ and denote by $p_0$ the uniform distribution on $\mathcal{Z}$, i.e. $p_0 = \frac{1}{N} \sum_{n=1}^N \delta_{z_n}$, where $\delta_{z_n}$ is the Dirac distribution supported at $z_n$. We may assume that there is a single equivalence class in $\argminf \mathbb{E}_{p_0}[\ell(Z, f)]$, otherwise we are done by Lemma \ref{lemma-tool-to-prove-LB}. Take $f^* \in \argminf \mathbb{E}_{p_0}[\ell(Z, f)]$. Since the game $\stdgame$ is not trivial, there exists $z_k$ in $\mathcal{Z}$ and $f^{**}$ in $\mathcal{F}$ such that $\ell(z_k, f^{**}) < \ell(z_k, f^*)$. Therefore, we can find $\epsilon >0$ small enough such that $(N-1) \epsilon < 1$ and:
$$
	(1 - (N-1) \epsilon) \ell(z_k, f^{**}) + \sum_{n \neq k} \epsilon \ell(z_n, f^{**}) < (1 - (N-1) \epsilon) \ell(z_k, f^*) + \sum_{n \neq k} \epsilon \ell(z_n, f^*).
$$
Define $p_1$ as the corresponding distribution, i.e. $p_1 = (1 - (N-1) \epsilon) \delta_{z_k} + \epsilon \sum_{n \neq k} \delta_{z_n}$. By construction, the equivalence class of $f^*$ is not in $\argminf \mathbb{E}_{p_1}[\ell(Z, f)]$. Once again, without loss of generality, we may assume that there is a single equivalence class  in $\argminf \mathbb{E}_{p_1}[\ell(Z, f)]$, otherwise we are done by Lemma \ref{lemma-tool-to-prove-LB}. Moreover, we can now redefine $f^{**}$ as a representative of the only equivalence class contained in $\argminf \mathbb{E}_{p_1}[\ell(Z, f)]$. We now move on to show that there must exist $\alpha \in (0, 1)$ such that there are at least two equivalence classes in $\argminf \mathbb{E}_{p_\alpha}[\ell(Z, f)]$, where the distribution $p_\alpha$ is defined as  $p_\alpha = (1 - \alpha) p_0 + \alpha p_1$. Observe that $\min_{f \in \mathcal{F}} \mathbb{E}_{p_\alpha}[\ell(Z, f)]$ can be written as the linear program:
	\begin{equation}
		\label{eq-proof-lp-OmegaSqrtDiscreteZ}
		\begin{aligned}
		& \min_{ q_1, \cdots, q_N, f } 
		& & q \cdot ( (1 - \alpha) x_0 + \alpha x_1) \\
		& \text{subject to}
		& & q = (q_1, \cdots, q_N) \\
		& 
		& & q_n \geq (C(z_n) f + c(z_n))^\tr x \quad \forall x \in \mathcal{X}(z_n), \forall n = 1, \cdots, N \\
		&
		& & f \in \mathcal{F}, q_1, \cdots, q_N \in \mathbb{R}
		\end{aligned}
	\end{equation}	
where $x_0$ and $x_1$ are vectors of size $N$ defined as follows: 
$$
	x_0 = \frac{1}{N} (1, \cdots, 1)
$$ 
and 
$$
	x_1 = (1 - (N-1) \epsilon) (0, \cdots, 0, 1, 0, \cdots)  + \epsilon (1, \cdots, 1, 0, 1, \cdots, 1),
$$ 
i.e. all the components of $x_1$ are equal to $\epsilon$ except for the $k$th one which is equal to $(1 - (N-1) \epsilon)$. We are interested in the function $\phi: \alpha \rightarrow \argminf \mathbb{E}_{p_\alpha}[\ell(Z, f)]$. For any $f \in \mathcal{F}$, define $I(f) = \{ \alpha \in [0,1] \; | \; f \in \phi(\alpha) \}$. Since $\alpha \rightarrow \mathbb{E}_{p_\alpha}[\ell(Z, f)]$ is linear in $\alpha$, $I(f) = \{ \alpha \in [0,1] \; | \; f \in \phi(\alpha) \}$ is a closed interval in $[0,1]$ for any $f \in \mathcal{F}$. Moreover, $\mathcal{F}$ being a polyhedron, the feasible set of \eqref{eq-proof-lp-OmegaSqrtDiscreteZ} is also a polyhedron, hence it has a finitely many extreme points. We denote by $\{f_1, \cdots, f_L\}$ the projection of the set of extreme points onto the $f$ coordinate. Since \eqref{eq-proof-lp-OmegaSqrtDiscreteZ} is a linear program, this shows that, for any $\alpha \in [0, 1]$, there exists $l \in \{1, \cdots, L\}$ such that $f_l \in \phi(\alpha)$. Therefore, we can write $[0, 1] = \cup_{l=1}^L I(f_l)$. We can further simplify this description by assuming that the $f_l$'s belong to different equivalence classes (because $I(f) = I(f')$ if $f$ is equivalent to $f'$). Now observe that if $I(f_l) \cap I(f_j) \neq \emptyset$ for all any $l \neq j \leq K$, then there are two classes of equivalence in $\argminf \mathbb{E}_{p_\alpha}[\ell(Z, f)]$ for any $\alpha \in I(f_l) \cap I(f_j)$ and we are done. Suppose by contradiction that we cannot find such a pair of indices. Because the only way to partition $[0,1]$ into $L < \infty$ non-overlapping closed intervals is to have $L = 1$, we get $[0, 1] = I(f_1)$. This implies that $f^*$ and $f^{**}$ belong to the same equivalence class, a contradiction.

\subsection{Alternative proof of Theorem \ref{lemma-OmegaSqrtPiecewiseLinear} by exhibiting an equalizing strategy when $\ell$ is linear}
Using Lemma \ref{lemma-LinearConvexGameIsEquivalent}, we can assume without loss of generality that $\mathcal{Z}$ is convex. When $\ell$ is linear, the procedure developed in the proof of Theorem \ref{lemma-OmegaSqrtPiecewiseLinear} boils down to finding a point $z \in \interior( \mathcal{Z} )$ such that $| \argminf z^\tr f | > 1$ and, with further examination, we can also guarantee that there exists $\epsilon > 0, e \in \mathbb{R}^n$ and $f_1, f_2 \in \argminf z ^\tr f$ such that $f_1 \in \argminf (z - x e) ^\tr f$ while $f_2 \notin \argminf (z - x e) ^\tr f$ for all $x \in (0, \epsilon]$ and symmetrically for $x \in [-\epsilon, 0)$. Consider a randomized opponent $Z_t = z + (\epsilon_t \epsilon) e$ for $(\epsilon_t)_{t=1, \cdots, T}$ i.i.d. Rademacher random variables. Then for any player's strategy:
$$
\mathbb{E}[r_T((Z_t)_{t=1, \cdots, T}, (f_t)_{t=1, \cdots, T})] = \sum_{t=1}^T \mathbb{E}[Z_t] ^\tr f_t - \mathbb{E}[ \inf\limits_{f \in \mathcal{F}} f ^\tr \sum_{t=1}^T Z_t].
$$
This yields:
$$
\mathbb{E}[r_T((Z_t)_{t=1, \cdots, T}, (f_t)_{t=1, \cdots, T})] = \sum_{t=1}^T z ^\tr f_t  - \mathbb{E}[ \inf\limits_{f \in \mathcal{F}} f ^\tr \sum_{t=1}^T Z_t].
$$
We can lower bound the last quantity by:
$$
\mathbb{E}[r_T((Z_t)_{t=1, \cdots, T}, (f_t)_{t=1, \cdots, T})] \geq T (z ^\tr f_1) - T \mathbb{E}[ \inf\limits_{f \in \mathcal{F}} f ^\tr (z + (\epsilon \cdot \frac{\sum_{t=1}^T \epsilon_t}{T})e)],
$$
as $f_1 \in \argminf z ^\tr f$, but we could have equivalently picked $f_2$ as $f_1 ^\tr z = f_2 ^\tr z$. Furthermore, as $|\frac{\sum_{t=1}^T \epsilon_t}{T}| \leq 1$, $f_1$ is optimal in the inner optimization problem when $\sum_{t=1}^T \epsilon_t\ \leq 0$ while $f_2$ is optimal when $\sum_{t=1}^T \epsilon_t \geq 0$. Hence:
\begin{multline*}
\mathbb{E}[r_T((Z_t)_{t=1, \cdots, T}, (f_t)_{t=1, \cdots, T})] \geq T (z ^\tr f_1) - T \mathbb{E}[ \; f_1 ^\tr (z + (\epsilon \cdot \frac{\sum_{t=1}^T \epsilon_t}{T})e) \cdot 1_{\sum_{t=1}^T \epsilon_t \leq 0} + \\ 
f_2 ^\tr (z + (\epsilon \cdot \frac{\sum_{t=1}^T \epsilon_t}{T})e) \cdot 1_{\sum_{t=1}^T \epsilon_t \geq 0} \; ].
\end{multline*}
Observe that the term $T (z ^\tr f_1)$ cancels out and we get:
$$
\mathbb{E}[r_T((Z_t)_{t=1, \cdots, T}, (f_t)_{t=1, \cdots, T})] \geq \frac{\mathbb{E}[|\sum_{t=1} \epsilon_t|]}{T} \cdot \epsilon \cdot (f_1 ^\tr e - f_2 ^\tr e).
$$
By Khintchine's inequality $\mathbb{E}[|\sum_{t=1}^T \epsilon_t|] \geq \frac{1}{\sqrt{2}} \sqrt{T}$. Moreover $f_1 ^\tr e - f_2 ^\tr e > 0$ because $f_2 \in \argminf (z + \epsilon e) ^\tr f$ while $f_1$ does not and $f_1 ^\tr z = f_2 ^\tr z$. We finally derive 
$$
\mathbb{E}[r_T((Z_t)_{t=1, \cdots, T}, (f_t)_{t=1, \cdots, T})] \geq \frac{(f_1 ^\tr e - f_2 ^\tr e)}{\sqrt{2}} \sqrt{T}.
$$ 
This enables us to conclude $R_T = \omegasqrt$ as this shows that for any player's strategy, there exists a sequence $z_1, \cdots, z_T$ such that 
$$
	r_T((z_t)_{t=1, \cdots, T}, (f_t)_{t=1, \cdots, T}) \geq \mathbb{E}[r_T((Z_t)_{t=1, \cdots, T}, (f_t)_{t=1, \cdots, T})].
$$

\subsection{Proof of Theorem \ref{lemma-OmegaSqrtLinearPolyhedron}}

Straightforward from Theorem \ref{lemma-OmegaSqrtPiecewiseLinear} since $\ell$ is jointly continuous.

\subsection{Proof of Lemma \ref{lemma-ZeroInInterior}}

Using Lemma \ref{lemma-LinearConvexGameIsEquivalent}, we can assume that $\mathcal{Z}$ is convex. Consider $f_1 \neq f_2 \in \mathcal{F}$ and define $e = \frac{f_1 - f_2}{\norm{f_1 - f_2}}$. Since $0 \in \interior(\mathcal{Z})$, there exists $\epsilon > 0$ such that $\epsilon e$ and $- \epsilon e$ are in $\mathcal{Z}$. We restrict the opponent's decision set by imposing that, at any round $t$, the opponent's move be $y_t \epsilon e$ for $y_t \in \tilde{\mathcal{Z}} = \{-1, 1\}$. Since $\ell(y_t \epsilon e, f)$ only depends on $f$ through the scalar product between $f$ and $e$, the player's decision set can equivalently be described by $\tilde{\mathcal{F}} = \{f^\tr e \; | \; f \in \mathcal{F} \}$ which is a closed interval (since $\mathcal{F}$ is convex and compact) and thus a polyhedron. Defining a new loss function as $\tilde{\ell}(y, f) = y \epsilon f$, we have:
$$
	R_T(\ell, \mathcal{Z}, \mathcal{F}) \geq R_T(\tilde{\ell}, \tilde{\mathcal{Z}}, \tilde{\mathcal{F}}).
$$
Observe that the game $(\tilde{\ell}, \tilde{\mathcal{Z}}, \tilde{\mathcal{F}})$ is linear and not trivial, otherwise there would exist $f^*$ such that $e^\tr f^* \leq e^\tr f_2$ and $-e^\tr f^* \leq - e^\tr f_1$ which would imply $\norm{e} = 0$. With Theorem \ref{lemma-OmegaSqrtLinearPolyhedron}, we conclude $R_T(\tilde{\ell}, \tilde{\mathcal{Z}}, \tilde{\mathcal{F}}) = \omegasqrt$ and thus $R_T(\ell, \mathcal{Z}, \mathcal{F}) = \omegasqrt$.

\subsection{Proof of Theorem \ref{lemma-ftl-yields-logT}}

A common inequality on the regret incurred for the FTL strategy is:
	$$
		r_T((z_t)_{t=1, \cdots, T}, (f_t)_{t=1, \cdots, T}) \leq \sum_{t=1}^T z_t^\tr (f_t - f_{t+1}),
	$$
	We use sensitivity analysis to control this last quantity. Specifically we show that the mapping $\phi: z \rightarrow \argmin_{f, F(f) = 0} z^\tr f$ is Lipschitz on $\mathcal{Z}$. Using this property:
	\begin{align*}
		r_T((z_t)_{t=1, \cdots, T}, (f_t)_{t=1, \cdots, T}) 
			& \leq \sum_{t=1}^T \norm{z_t} \norm{f_t - f_{t+1}} \\
			& = O(\sum_{t=1}^T \norm{\frac{1}{t-1} \sum_{\tau = 1}^{t-1} z_\tau - \frac{1}{t} \sum_{\tau = 1}^{t} z_\tau }) \\
			& = O(\sum_{t=1}^T \norm{\frac{1}{t (t-1)} \sum_{\tau = 1}^{t-1} z_\tau - \frac{1}{t} z_t }) \\
			& = O(\sum_{t=1}^T \frac{1}{t (t-1)} \norm{\sum_{\tau = 1}^{t-1} z_\tau} +  \frac{1}{t} \norm{z_t} ) \\
			& = O(\sum_{t=1}^T \frac{1}{t}) \\
			& = O( \log(T) ),
	\end{align*}
	since $\mathcal{Z}$ is compact.	
	\\
	We now move on to show that $\phi$ is Lipschitz. As $\conv(\mathcal{Z})$ is closed and convex, we can strictly separate $0$ from $\conv(\mathcal{Z})$. Hence, there exists $a \neq 0 \in \mathbb{R}^d$ and $c > 0$ such that $a \cdot z > c$, $\forall z \in \mathcal{Z}$. We get $\| z \| \geq \frac{c}{\|a\|} > 0$ $\forall z \in \mathcal{Z}$. Let us use the shorthand $C = \frac{c}{\|a\|}$. Take $(z_1, z_2) \in \mathcal{Z}^2$ and $(f(z_1), f(z_2)) \in \phi(z_1) \times \phi(z_2)$. Observe that the constraint qualifications are automatically satisfied at $f(z_1)$ and $f(z_2)$ as $\nabla F$ cannot vanish on $\{ f \; | \; F(f) = 0\}$ since $F$ does cannot attain its minimum on this set ($\mathcal{F}$ is assumed to contain at least two points). Hence, there exist $\lambda_1, \lambda_2 \geq 0$ such that $z_1 + \lambda_1 \nabla F(f(z_1)) = 0$ and $z_2 + \lambda_2 \nabla F(f(z_2)) = 0$. As $z_1, z_2 \neq 0$, we must have $\lambda_1, \lambda_2 \neq 0$. We obtain $\nabla F(f_1) = - \frac{1}{\lambda_1} z_1$ and $\nabla F(f(z_2)) = - \frac{1}{\lambda_2} z_2$. Since $F$ is strongly convex, there exists $\beta > 0$ such that:
		$$
			( \nabla F(f') - \nabla F(f'') ) ^\tr (f' - f'') \geq \beta \norm{ f' - f'' }^2,
		$$
		for all $f', f'' \in \mathcal{F}$. Applying this inequality for $f' = f(z_1)$ and $f'' = f(z_2)$, we obtain:
		$$
			( \frac{1}{\lambda_2} z_2 - \frac{1}{\lambda_1} z_1 ) ^\tr (f(z_1) - f(z_2)) \geq \beta \norm{f(z_1) - f(z_2)}^2.
		$$
		We can break down the last expression in two pieces:
		$$
			\frac{1}{\lambda_2} z_2 ^\tr (f(z_1) - f(z_2))  + \frac{1}{\lambda_1} z_1 ^\tr (f(z_2) - f(z_1)) \geq \beta \norm{f(z_1) - f(z_2) }^2.
		$$		
		Observe that $z_2 ^\tr (f(z_1) - f(z_2)) \geq 0$ since both $f(z_1)$ and $f(z_2)$ belong to $\mathcal{F}$ and since $f(z_2)$ is the minimizer of $z_2 ^\tr f$ for $f$ ranging in $\mathcal{F}$. Symmetrically, $z_1 ^\tr (f(z_2) - f(z_1)) \geq 0$. Note that $\frac{1}{\lambda_1} = \frac{1}{|\lambda_1|}= \frac{\| \nabla F(f(z_1)) \|}{\| z_1 \|}$. As $\nabla F$ is continuous and $\mathcal{F}$ is compact, there exists $K > 0$ such that $\| \nabla F(f) \| \leq K$ for any $f \in \mathcal{F}$. Hence, we get $\frac{1}{\lambda_1} \leq \frac{K}{C}$ and the same inequality holds for $\lambda_2$. Plugging this upper bound back into the last inequality yields:
		$$
			\frac{K}{C} ( z_2 - z_1 ) ^\tr (f(z_1) - f(z_2)) \geq \beta \| f(z_1) - f(z_2) \|^2.
		$$		
		Using the Cauchy-Schwarz inequality and simplifying on both sides by $\| f(z_2) - f(z_1) \|$ yields:
		$$
			\frac{K}{\beta C} \| z_2 - z_1 \| \geq \| f(z_1) - f(z_2) \|,
		$$	
		i.e. $\frac{K}{\beta C} \| z_2 - z_1 \| \geq \| \phi(z_1) - \phi(z_2) \|$.
	
\subsection{Proof of Lemma \ref{lemma-NotEnoughIID}}

Using Lemma \ref{lemma-LinearConvexGameIsEquivalent}, we can assume that $\mathcal{Z}$ is convex. Since $\mathcal{Z}$ is compact and convex and since $0 \notin \mathcal{Z}$, we can strictly separate $0$ from $\mathcal{Z}$ and find $z^* \neq 0$ such that $\mathcal{Z} \subseteq B_2(z^*, \alpha \norm{z^*})$ with $\alpha < 1$. By rescaling $\mathcal{Z}$, we can assume that $\alpha \norm{z^*} = 1$ and $\norm{z^*} > 1$. In the sequel, $\sigma_{t-1}$ serves as a shorthand for $\sigma(Z_1, \cdots, Z_{t-1})$. We prove more generally that, for any choice of random variables $(Z_1, \cdots, Z_T)$ such that $\mathbb{E}[Z_t | \sigma_{t-1} ]$ is constant almost surely, the lower bound on regret derived from Theorem \ref{lemma-minimax-reformulation} is $O(1)$. Write $Z_t = z^* + W_t$ and $\mathbb{E}[W_t | \sigma_{t-1}] = c_t$ with $\| W_t \| \leq 1$ and $\| c_t \| \leq 1$. Define $w^* = T \cdot z^* + \sum_{t=1}^T c_t$. Observe that $\| w^* \| \geq T \cdot \| z^* \| - \| \sum_{t=1}^T c_t \| \geq T \cdot (\| z^* \| -1) > 0$. Write $W_t = X_t \frac{w^*}{\| w^* \|} + \tilde{W}_t + c_t$ with $\tilde{W}_t^\tr w^* = 0$. Projecting down the equality $ \mathbb{E}[W_t - c_t| \sigma_{t-1}] = 0$ onto $w^*$, we get $\mathbb{E}[X_t | \sigma_{t-1}] = 0$ and $\mathbb{E}[\tilde{W}_t | \sigma_{t-1}] = 0$. The bound that results from an application of Theorem \ref{lemma-minimax-reformulation} is:
$$
	R_T \geq \mathbb{E}[ \| w^* + \sum_{t=1}^T W_t - c_t \|] - \sum_{t=1}^T \| z^* + c_t \|].
$$
We now focus on finding an upper bound on the right-hand side. Expanding the first term yields: 
$$
	\| w^* + \sum_{t=1}^T W_t - c_t \| = \sqrt{ (1 + \sum_{t=1}^T \frac{X_t}{\| w^* \| })^2 \cdot \| w^* \|^2 + \| \sum_{t=1}^T \tilde{W}_t \|^2 }.
$$ 
By concavity of the squared root function:
$$
	\mathbb{E}[ \| w^* + \sum_{t=1}^T W_t - c_t \| ] \leq \sqrt{ \| w^* \|^2 \cdot \mathbb{E}[ (1 + \sum_{t=1}^T \frac{X_t}{\| w^* \| })^2 ] + \mathbb{E}[ \| \sum_{t=1}^T \tilde{W}_t \|^2 ] }.
$$
We expand the two inner terms: 
$$ 
\mathbb{E}[ (1 + \sum_{t=1}^T \frac{X_t}{\| w^* \| })^2 ] = 1 +  2 \sum_{t=1}^T \frac{ \mathbb{E}[X_t]}{\| w^* \| } + \frac{1}{\| w^* \|^2} \cdot \mathbb{E}[(\sum_{t=1}^T X_t)^2]. 
$$ 
Looking at each term individually, we have $\mathbb{E}[X_t] = \mathbb{E}[ \mathbb{E}[X_t | \sigma_{t-1} ] = 0$ and $\mathbb{E}[(\sum_{t=1}^T X_t)^2] = \mathbb{E}[(\sum_{t=1}^{T-1} X_t)^2] + 2  \mathbb{E}[X_T \cdot (\sum_{t=1}^{T-1} X_t)] + \mathbb{E}[X_T^2]$, yet $\mathbb{E}[X_T \cdot (\sum_{t=1}^{T-1} X_t)] = \mathbb{E}[ \mathbb{E}[ X_T | \sigma_{T-1}] \cdot (\sum_{t=1}^{T-1} X_t)] = 0$. Hence, $ \mathbb{E}[ (1 + \sum_{t=1}^T \frac{X_t}{\| w^* \| })^2 ] = 1 + \frac{\mathbb{E}[ \sum_{t=1}^T X_t^2 ] }{\| w^* \|^2} $. Similarly $\mathbb{E}[ \| \sum_{t=1}^T \tilde{W}_t \|^2 ] = \sum_{t=1}^T \mathbb{E}[ \| \tilde{W}_t \|^2 ]$. We obtain:
$$
	\mathbb{E}[ \| w^* + \sum_{t=1}^T W_t - c_t \| ] \leq \sqrt{ \| w^* \|^2 + \sum_{t=1}^T \mathbb{E}[ X_t^2 + \| \tilde{W}_t \|^2 ]  }.
$$
Remark that $\| W_t - c_t \| \leq \| W_t \| + \| c_t \| \leq 2$. Hence, $ X_t^2 + \| \tilde{W}_t \|^2 \leq 2 $. We obtain: 
$$
	\mathbb{E}[ \| w^* + \sum_{t=1}^T W_t - c_t \| ] \leq \sqrt{ \| w^* \|^2 + 2 T }.
$$
We have $\sqrt{ \| w^* \|^2 + 2  T} = \| w^* \| \cdot \sqrt{1 + \frac{2  T}{\| w^* \|^2}} \leq \| w^* \| + \frac{T}{\| w^* \|}$ for $T$ big enough as $\| w^* \| \geq T \cdot (\| z^*\| - 1)$. Yet $\| w^* \| = \| \sum_{t=1}^T z^* + c_t \| \leq \sum_{t=1}^T \| z^* + c_t \|$. Hence, the lower bound derived is:
$$
\mathbb{E}[ \| w^* + \sum_{t=1}^T W_t - c_t \|] - \sum_{t=1}^T \| z^* + c_t \| \leq \frac{T}{\| w^* \|} \leq \frac{1}{\| z^* \| - 1} = O(1).
$$

\subsection{Proof of Theorem \ref{lemma-NonIIDLowerBound}}
Since $\mathcal{Z}$ has non-empty interior, we can find $z^* \neq 0$ and $\alpha \in (0, \frac{1}{32}]$ such that $B_2(z^*, \alpha \norm{z^*}) \subseteq \mathcal{Z}$. Define $e$ as a unit vector orthogonal to $z^*$ and $\tilde{\mathcal{Z}} = \{ z \; | z = z^* + (w \alpha \norm{z^*}) e, |w| \leq 1 \}$. Since $\tilde{\mathcal{Z}} \subseteq \mathcal{Z}$, we have:
$$
	R_T(\ell, \mathcal{Z}, \mathcal{F}) \geq R_T(\ell, \tilde{\mathcal{Z}}, \mathcal{F}), 
$$
and we can focus on developing a $\Omega(\log(T))$ lower bound on regret for the game $(\ell, \tilde{\mathcal{Z}}, \mathcal{F})$. Using the minimax reformulation of Theorem \ref{lemma-minimax-reformulation}, we have:
\begin{align*}
	& R_T(\ell, \tilde{\mathcal{Z}}, \mathcal{F}) \\
	& = \sup\limits_{p} \mathbb{E} \left[ \; - \sum_{t=1}^T \norm{z^* + (\alpha \norm{z^*} \mathbb{E}[ W_t | W_1, \cdots, W_{t-1} ])e } + \norm{T z^* + (\alpha \norm{z^*} \sum_{t=1}^T W_t) e } \right] \\
	& = \sup\limits_{p} \mathbb{E} \left[ \; - \sum_{t=1}^T \sqrt{ \norm{z^*}^2 + (\alpha \norm{z^*} \mathbb{E}[ W_t | W_1, \cdots, W_{t-1} ])^2 } + \sqrt{ T^2 \norm{z^*}^2 + (\alpha \norm{z^*} \sum_{t=1}^T W_t)^2  } \right]
\end{align*}
where the supremum is taken over the distribution $p$ of the random variable $(W_1, \cdots, W_T)$ in $[- 1, 1 ]^T$. Rearranging this expression yields:

\begin{alignat*}{2}
		R_T(\ell, \tilde{\mathcal{Z}}, \mathcal{F}) 
			& = \norm{z^*} \sup\limits_{p} && \mathbb{E} \left[ \; T \sqrt{ 1 + (\alpha \frac{\sum_{t=1}^T W_t}{T})^2  } - \sum_{t=1}^T \sqrt{ 1+ (\alpha \mathbb{E}[ W_t | W_1, \cdots, W_{t-1} ])^2 }  \right] \\
			& =  \norm{z^*} \sup\limits_{p} \{ && \mathbb{E} [ \; T ( 1 + \sum_{n=1}^\infty {\frac{1}{2} \choose n } \alpha^{2n} (\frac{\sum_{t=1}^T W_t}{T})^{2n} ) \\
			& && - \sum_{t=1}^T ( 1 + \sum_{n=1}^\infty {\frac{1}{2} \choose n } \alpha^{2n} \mathbb{E}[ W_t | W_1, \cdots, W_{t-1} ]^{2n} )  ] \} \\
			& =  \norm{z^*} \sup\limits_{p} \{ && \frac{\alpha^2}{2} \mathbb{E} \left[ \; \frac{ (\sum_{t=1}^T W_t)^2 }{T} - \sum_{t=1}^T \mathbb{E}[ W_t | W_1, \cdots, W_{t-1} ]^2 \right] \\
			& && + \sum_{n=2}^\infty {\frac{1}{2} \choose n } \alpha^{2n} \mathbb{E} \left[\frac{(\sum_{t=1}^T W_t)^{2n}}{ T^{2n -1} } - \sum_{t=1}^T \mathbb{E}[ W_t | W_1, \cdots, W_{t-1} ]^{2n} \right] \}, \\
\end{alignat*}
where the second equality results from a series expansion (valid since $(\alpha \mathbb{E}[ W_t | W_1, \cdots, W_{t-1} ])^2, (\alpha \frac{\sum_{t=1}^T W_t}{T})^2 \leq \alpha^2 < 1$) and the third inequality is derived from Fubini, observing that:
$$
	\sum_{n=1}^\infty |{\frac{1}{2} \choose n }| \alpha^{2n} \mathbb{E}[(\frac{\sum_{t=1}^T W_t}{T})^{2n}] \leq \sum_{n=1}^\infty \alpha^{2n} = \frac{1}{ 1 - \alpha^2} < \infty
$$
and similarly:
$$
	\sum_{n=1}^\infty |{\frac{1}{2} \choose n }| \alpha^{2n} \mathbb{E}[\mathbb{E}[ W_t | W_1, \cdots, W_{t-1} ]^{2n} ] \leq \sum_{n=1}^\infty \alpha^{2n} = \frac{1}{ 1 - \alpha^2} < \infty.
$$
Interestingly, the first-order term of this series expansion, i.e. 
$$
	\mathbb{E} \left[ \; \frac{ (\sum_{t=1}^T W_t)^2 }{T} - \sum_{t=1}^T \mathbb{E}[ W_t | W_1, \cdots, W_{t-1} ]^2 \right],
$$
is precisely the expression of the minimax regret for the game $(\ell(z, f) = (z - f)^2, [-1, 1], [-1, 1])$ which is known to have optimal regret $\Theta(\log(T))$, see Section 7.3 of \cite{abernethy2009stochastic}. This motivates the introduction of the probability distribution $p$ used in \cite{abernethy2009stochastic} to establish the $\Omega(\log(T))$ lower bound. Specifically, we use the conditional distributions:
$$
	p_t(W_t = w | W_1, \cdots, W_{t-1}) =  \left\{ \begin{array}{rl}
						 						\frac{1 + c_t W_{1:t-1}}{2} &\mbox{ if $w=1$} \\
					  						    \frac{1 - c_t W_{1:t-1}}{2} &\mbox{ if $w=-1$}
				  							\end{array} 
		  									\right. \quad t = 2, \cdots, T
$$
where $W_{1:t-1} = \sum_{\tau=1}^{t-1} W_\tau$ and the sequence $(c_t)_{t=1, \cdots, T}$ is recursively defined as:
\begin{align*}
	c_T & =  \frac{1}{T} \\
	c_{t-1} & =  c_t + c_t^2 \quad t = T, \cdots, 2. 
\end{align*}
Together with $W_1$ taken as a Rademacher random variable, these conditional distributions define a joint distribution $p$ as it can be shown that $c_t \in [0, \frac{1}{t}]$. \citet{abernethy2009stochastic} show that: 
\begin{equation}
	\label{eq-proof-log-T-from-Abernethy}
	\mathbb{E} \left[ \; \frac{ (\sum_{t=1}^T W_t)^2 }{T} - \sum_{t=1}^T \mathbb{E}[ W_t | W_1, \cdots, W_{t-1} ]^2 \right] = \log(T) + O(\log\log(T)).
\end{equation}
Hence, it remains to control the terms of order $n \geq 2$ in the series expansion. First observe that, by definition:
\begin{align*}
	\mathbb{E}[W_{1:T}^{2n}] 
		& = \mathbb{E}[ \mathbb{E}[W_{1:T}^{2n} | W_1, \cdots, W_{T-1}]] \\
		& = \mathbb{E}[ \frac{1 + c_t W_{1:T-1}}{2} ( W_{1:T-1} + 1)^{2n} + \frac{1 - c_t W_{1:T-1}}{2} ( W_{1:T-1} + 1)^{2n} ] \\
		& = 1 + \sum_{k=1}^{n} ({2n \choose 2k} + {2n \choose 2k-1} c_T) \mathbb{E}[ (W_{1:T-1})^{2k} ],
\end{align*}
which implies that:
\begin{equation}
\label{eq-proof-log-T-lowerbound-recineq-W}
\begin{aligned}
	| c_T^{2n-1} \mathbb{E}[W_{1:T}^{2n}] - (c_T^{2n-1} + 2 n c_T^{2n}) \mathbb{E}[W_{1:T-1}^{2n}] |
		& \leq  {2n \choose 2 (n-1)} c_T + 2 \sum_{k=0}^{n} {2n \choose 2k} c_T^2 \\
		& \leq 2 n^2 c_T + 2 4^n c_T^2,
\end{aligned}
\end{equation}
since $c_T |W_{1:T-1}| \leq 1$. Additionally, we have, using the recursive definition of the sequence $(c_t)_{t=1, \cdots, T}$:
\begin{align*}
	c_{T-1}^{2n-1} 
		& = (c_T + c_T^{2} )^{2n-1} \\
		& = \sum_{k=0}^{2n-1} {2n-1 \choose k} c_T^{2n-1+k},
\end{align*}
which implies:
\begin{equation}
\label{eq-proof-log-T-lowerbound-recineq-C}
	|c_{T-1}^{2n-1}  - (c_T^{2n-1} + (2 n-1) c_T^{2n})| \leq 2 n^2 c_T^{2n+1} + 4^n c_T^{2n+2}.
\end{equation}
Using $\mathbb{E}[ W_T | W_1, \cdots, W_{T-1} ] = c_T W_{1:T-1}$, we get:
\begin{align*}
	| \mathbb{E} [ c_T^{2n-1} W_{1:T}^{2n} - & \sum_{t=1}^T \mathbb{E}[ W_t | W_1, \cdots, W_{t-1} ]^{2n} ] | \\
		& \leq 	| \mathbb{E}[(c_T^{2n-1} + 2 (n-1) c_T^{2n}) W_{1:T-1}^{2n}] - \sum_{t=1}^{T-1} \mathbb{E}[ W_t | W_1, \cdots, W_{t-1} ]^{2n} ] | \\	
		& + 2 n^2 c_T + 2 4^n c_T^2  \\
		& \leq 	| \mathbb{E}[c_{T-1}^{2n-1} W_{1:T-1}^{2n} - \sum_{t=1}^{T-1} \mathbb{E}[ W_t | W_1, \cdots, W_{t-1} ]^{2n} ] | \\	
		& + (2 n^2 c_T^{2n+1} + 4^n c_T^{2n+2}) \mathbb{E}[W_{1:T-1}^{2n}] + 2 n^2 c_T + 2 4^n c_T^2  \\
		& \leq 4 n^2 c_T + 3 4^n c_T^2  \\
		& \leq 4 n^2 \frac{1}{T} + 3 4^n \frac{1}{T^2},
\end{align*}
where the first (resp. second) inequality is obtained by applying \eqref{eq-proof-log-T-lowerbound-recineq-W} (resp. \eqref{eq-proof-log-T-lowerbound-recineq-C}) and the fifth inequality is derived using $c_T \in [0, \frac{1}{T}]$ and $|W_{1:T-1}| \leq T - 1$. By induction on $t$, we get:
\begin{align*}
	| \mathbb{E} \left[ c_T^{2n-1} W_{1:T}^{2n} - \sum_{t=1}^T \mathbb{E}[ W_t | W_1, \cdots, W_{t-1} ]^{2n} \right] | 
		& \leq 4 n^2 \sum_{t=1}^T \frac{1}{t} + 3 4^n \sum_{t=1}^T \frac{1}{t^2}  \\
		& \leq 4 n^2 \log(T) + 4^n \frac{\pi^2}{2}.
\end{align*}
Bringing everything together, we derive:
\begin{align*}
	|\sum_{n=2}^\infty {\frac{1}{2} \choose n } \alpha^{2n} & \mathbb{E} \left[\frac{(\sum_{t=1}^T W_t)^{2n}}{ T^{2n -1} } - \sum_{t=1}^T \mathbb{E}[ W_t | W_1, \cdots, W_{t-1} ]^{2n} \right] | \\
	& \leq 4 (\sum_{n=2}^\infty {\frac{1}{2} \choose n } \alpha^{2n} n^2) \log(T) \\
	& + (\sum_{n=2}^\infty {\frac{1}{2} \choose n } (2 \alpha)^{2n}) \frac{\pi^2}{2} \\
	& \leq 8 \alpha^4 (\sum_{n=2}^\infty n (n-1) (\alpha^2)^{n-2}) \log(T) \\
	& + (\sum_{n=0}^\infty (2 \alpha)^{2n}) \frac{\pi^2}{2} \\
	& \leq 8 \frac{\alpha^4}{(1-\alpha^2)^3} \log(T) + \frac{\pi^2}{2 ( 1 - 2 \alpha )}\\
	& \leq 8 \frac{\alpha^4}{(1-\alpha^2)^3} \log(T) + \pi^2,
\end{align*}
since $\alpha \leq \frac{1}{4}$. Using \eqref{eq-proof-log-T-from-Abernethy}, we conclude that:
$$
	R_T(\ell, \tilde{\mathcal{Z}}, \mathcal{F}) \geq \frac{\norm{z^*} \alpha^2}{2} (1 - 16 \frac{\alpha^2}{(1-\alpha^2)^3}) \log(T) + O(\log\log(T)),
$$
which implies that $R_T(\ell, \tilde{\mathcal{Z}}, \mathcal{F}) = \Omega(\log(T))$ as $\alpha^2 (1 - 16 \frac{\alpha^2}{(1-\alpha^2)^3}) > 0$ for $\alpha \in (0, \frac{1}{32}]$.

\subsection{Proof of Lemma \ref{lemma-UniConvex-UpperBound}}
The proof is along the same lines as for Theorem \ref{lemma-ftl-yields-logT}. We start with the same inequality:
	$$
		r_T((z_t)_{t=1, \cdots, T}, (f_t)_{t=1, \cdots, T}) \leq \sum_{t=1}^T z_t^\tr (f_t - f_{t+1}),
	$$
	and use sensitivity analysis to control this last quantity. Specifically, we show that the mapping $\phi: z \rightarrow \argmin_{f \in \mathcal{F}} z^\tr f$ is $\frac{1}{q-1}$-H\"{o}lder continuous on $\mathcal{Z}$, i.e. there exists $c > 0$ such that:
	$$
		\norm{ \phi(z_1) - \phi(z_2) }_2 \leq c \norm{ z_1 - z_2 }_2^{\frac{1}{q-1}} \quad \forall (z_1, z_2) \in \mathcal{Z}^2.
	$$	
	Using this property, we get:
	\begin{align*}
		r_T((z_t)_{t=1, \cdots, T}, (f_t)_{t=1, \cdots, T}) 
			& \leq \sum_{t=1}^T \norm{z_t}_2 \norm{f_t - f_{t+1}}_2 \\
			& = O(\sum_{t=1}^T \norm{\frac{1}{t-1} \sum_{\tau = 1}^{t-1} z_\tau - \frac{1}{t} \sum_{\tau = 1}^{t} z_\tau }_2^{\frac{1}{q-1}}  ) \\
			& = O(\sum_{t=1}^T \norm{\frac{1}{t (t-1)} \sum_{\tau = 1}^{t-1} z_\tau - \frac{1}{t} z_t }_2^{\frac{1}{q-1}}   ) \\
			& = O(\sum_{t=1}^T (\frac{1}{t (t-1)} \norm{\sum_{\tau = 1}^{t-1} z_\tau}_2 +  \frac{1}{t} \norm{z_t}_2)^{\frac{1}{q-1}} ) \\
			& = O(\sum_{t=1}^T \frac{1}{t^{\frac{1}{q-1}}}),
	\end{align*}
	from which we derive that $r_T((z_t)_{t=1, \cdots, T}, (f_t)_{t=1, \cdots, T}) = O( \log(T) )$ if $q=2$ and $r_T((z_t)_{t=1, \cdots, T}, (f_t)_{t=1, \cdots, T}) = O( T^{\frac{q-2}{q-1}} )$ if $q \in (2, 3]$.
	\\
	We now move on to show that $\phi$ is $\frac{1}{q-1}$-H\"{o}lder continuous. Just like in Theorem \ref{lemma-ftl-yields-logT}, we can find $A > 0$ such that $\norm{z}_2 \geq A$ for all $z \in \mathcal{Z}$. Take $(z_1, z_2) \in \mathcal{Z}^2$ and $(f_1, f_2) \in \phi(z_1) \times \phi(z_2)$. Since we are optimizing a linear function, we may assume, without loss of generality, that $C=1$ and $f_1$ and $f_2$ lie on the boundary of $\mathcal{F}$, i.e. $\norm{f_1}_\mathcal{F} = \norm{f_2}_\mathcal{F} = 1$. By definition, we have:
	$$
		\norm{ \frac{f_1 + f_2}{2} }_\mathcal{F} \leq 1 - \delta_\mathcal{F}( \norm{f_1 - f_2}_\mathcal{F} ).
	$$
	As a consequence, we have:
	$$
		\frac{f_1 + f_2}{2} - \delta_\mathcal{F}( \norm{f_1 - f_2}_\mathcal{F} )  \frac{z_2}{ \norm{z_2}_\mathcal{F} } \in \mathcal{F}.
	$$
	We get:
	$$
		z_2^\tr (\frac{f_1 + f_2}{2} - \delta_\mathcal{F}( \norm{f_1 - f_2}_\mathcal{F} )  \frac{z_2}{ \norm{z_2}_\mathcal{F} } ) \geq \inf_{f \in \mathcal{F}} z_2^\tr f = z_2^\tr f_2.
	$$
	Rearranging this last inequality yields:
	$$	
		z_2^\tr \frac{f_1 - f_2}{2} \geq \frac{\norm{z_2}_2^2}{\norm{z_2}_\mathcal{F}} \delta_\mathcal{F}( \norm{f_1 - f_2}_\mathcal{F} ),
	$$
	which implies that:
	$$	
		z_2^\tr \frac{f_1 - f_2}{2} \geq K \norm{f_1 - f_2}_2^q,
	$$
	for some $K > 0$ independent of $z_1$ and $z_2$ since $\mathcal{Z}$ is compact, $\norm{z_2}_2 \geq A > 0$, $\norm{\;}_\mathcal{F}$ is $q$-uniformly convex, and by the equivalence of norms in finite dimensions. By optimality of $f_1$, we also have $z_1^\tr \frac{f_2 - f_1}{2} \geq 0$. Summing up the last two inequalities, we get:
	$$
		(z_2 - z_1)^\tr \frac{f_1 - f_2}{2} \geq K \norm{f_1 - f_2}_2^q,
	$$
	and (by Cauchy-Schwartz):
	$$
		\norm{z_2 - z_1}_2 \geq 2 K \norm{f_1 - f_2}_2^{q-1},
	$$
	which concludes the proof.

\end{document}